\documentclass[11pt]{article}

\usepackage[margin=1in]{geometry}
\usepackage{amsmath,amssymb,amsthm,bm}
\usepackage{booktabs}
\usepackage{graphicx}
\usepackage{float}
\usepackage{placeins}
\usepackage{xcolor}
\usepackage[numbers,sort&compress]{natbib}
\usepackage{microtype}
\usepackage{array}
\usepackage{enumitem}
\usepackage{hyperref}
\hypersetup{
  hidelinks,
  pdftitle={What a Data Release Must Preserve for Off-Policy Evaluation},
  pdfauthor={Kyungbok Lee, Yating Zou, Michael R. Kosorok}
}

\newcounter{algorithm}
\newenvironment{algorithmbox}[1]{%
  \refstepcounter{algorithm}%
  \par\medskip\noindent
  \begin{minipage}{\linewidth}
  \hrule\vspace{2pt}
  \textbf{Algorithm \thealgorithm.} #1\par
  \vspace{2pt}\hrule\vspace{4pt}
  \small
  \setlength{\abovedisplayskip}{4pt plus 1pt minus 1pt}%
  \setlength{\belowdisplayskip}{4pt plus 1pt minus 1pt}%
  \setlength{\abovedisplayshortskip}{2pt plus 1pt}%
  \setlength{\belowdisplayshortskip}{3pt plus 1pt minus 1pt}%
}{%
  \vspace{2pt}\hrule
  \end{minipage}
  \par\medskip
}

\newcommand{\cH}{\mathcal{H}}

\newcommand{\E}{\mathbb{E}}
\newcommand{\Pp}{\mathbb{P}}
\newcommand{\ind}{\mathbf{1}}

\newtheorem{theorem}{Theorem}[section]
\newtheorem{proposition}[theorem]{Proposition}
\newtheorem{lemma}[theorem]{Lemma}
\newtheorem{corollary}[theorem]{Corollary}

\title{What a Data Release Must Preserve for Off-Policy Evaluation}
\author{Kyungbok Lee \qquad Yating Zou \qquad Michael R. Kosorok\\
Department of Biostatistics, Gillings School of Global Public Health\\
University of North Carolina at Chapel Hill}
\date{}

\begin{document}
\maketitle

\begin{abstract}
In off-policy evaluation, merging decision-time histories that matter
differently for a specified policy value or comparison can make that evaluation
unidentified, even with observed actions and outcomes and known behavior and
target policies. We characterize which distinctions must be preserved:
identification holds if and only if the action-specific evaluation weights are
constant within each released record, and the coarsest exact release groups
histories by these weights. For normalized outcomes, the largest sharp
width induced by the release equals a behavior-weighted within-record radius of
these weights.

The same radius becomes a release-design objective when each record may be
refined into at most $m$ categories. The optimum has an exact $m$-center
characterization. When behavior does not vary within a released record, we give
a finite-data repair algorithm whose maximum evaluation-weight radius is at
most twice the finite-data optimum.

For a fixed action--outcome distribution within a released record, a distinct
sharp-identification problem remains. Under the same condition, the
action-specific outcome distributions are identified, and optimizing over their
couplings gives the sharp interval. We construct an AIPW envelope containing
this sharp interval; it matches the interval for all such outcome distributions
exactly when the extreme within-record action-coefficient vectors take at most
two values in each action coordinate. In 24 KuaiRec tasks, the repair algorithm
yields a $93\%$ median reduction in mean pairwise sharp width at $m=64$
relative to the unrepaired release.

\end{abstract}

\section{Introduction}

In Yahoo!'s Today Module, 1,193 user features were reduced to six dimensions
for contextual-bandit recommendation. The public release provides only the
reduced user features as context and supports offline policy evaluation
\citep{li2010contextual,li2011unbiased,wang2016hidden}. At a single decision
point, this kind of interface raises an OPE question before estimator choice:
actions may depend on a detailed pre-action history $H$, while the evaluator
sees only a deterministic release $G=R(H)$. Which distinctions in $H$ must
survive in $G$ for specified policy values and comparisons to remain
identified, even when actions and outcomes are observed and the behavior and
target policies on $H$ are known?

The issue is not how much information the release retains, but whether it preserves the distinctions among histories needed to identify the evaluations of interest. An \emph{evaluation contract} specifies the policy values and comparisons the release should support.  
Two policy values may each be ambiguous even when their comparison is identified.
Once a needed distinction has been lost in the release, no amount of additional samples can recover it.  
Whether the release supports the contract is therefore a structural identification question.

This perspective separates the analysis into two layers.
Sections~\ref{sec:identification}--\ref{sec:adequacy} concern the release
itself: the distinctions it must preserve, the worst-case ambiguity induced by
lost distinctions, and finite-category refinement.
Sections~\ref{sec:transport}--\ref{sec:contract-geometry-bridge} instead ask
what ambiguity remains for a particular action--outcome law within a released
record. They characterize the sharp interval and study the AIPW envelope
relative to that interval. The main results within these layers are:
\begin{itemize}[leftmargin=1.25em,itemsep=2pt,topsep=3pt]
\item \textbf{Exactness and worst-case ambiguity.}
We show that the values and comparisons in the contract are identified exactly
when their action-specific evaluation weights are constant within each released
record. The resulting weight classes define the coarsest exact release. For
$[0,1]$ outcomes, the largest sharp width the release can induce equals a
behavior-weighted within-record radius of those weights.

\item \textbf{Finite-category refinement.}
The same radius becomes the design objective when each released record may be
refined into at most $m$ categories. The optimal refinement has an exact
$m$-center characterization. When behavior does not vary within a released
record, we give a finite-data repair algorithm whose maximum evaluation-weight
radius is at most twice the finite-data optimum.

\item \textbf{Sharp identification.}
For a fixed action--outcome law within a released record, each action-specific
outcome distribution is identified exactly when behavior does not vary within
the record. Under this condition, optimizing over their couplings gives the
sharp interval. For a policy comparison, the sharp interval obtained directly
can be narrower than one derived from separate sharp intervals for the
component policy values.

\item \textbf{AIPW envelope and sharpness.}
Under the same within-record behavior condition, we construct an AIPW envelope
containing the fixed-law sharp interval and derive the exact slack at
corresponding endpoints. The envelope matches the
sharp interval for every collection of integrable action-specific outcome
distributions exactly when each action coordinate
takes at most two values across the extreme points of the within-record
action-coefficient set.
\end{itemize}

State abstraction, quotient representations, and target-oriented compression
study how information can be reduced while retaining what matters for
downstream tasks
\citep{hao2024abstracted,chaudhari2024arp,xie2026quotient,chang2026target}.
For this release problem, the criterion is whether the released law identifies
every value and comparison in the evaluation contract. When exactness fails,
the resulting ambiguity is a partial-identification problem
\citep{khan2024overlap,guo2022noisy}. Its worst-case magnitude has an
optimal-recovery radius interpretation \citep{donoho1994optimal}. For a fixed
action--outcome law within a released record, the sharp-identification problem
connects to multi-marginal transport \citep{gao2025multimarginal}. The AIPW
envelope builds on the classical doubly robust OPE score
\citep{dudik2011doubly}. We study the envelope's slack relative to the sharp
interval. Appendix~\ref{app:extended-positioning} gives further connections and
scope boundaries.

The empirical studies keep worst-case ambiguity, sharp identification, and
AIPW slack distinct. In KuaiRec, refinements are first compared by worst-case
ambiguity without using held-out outcomes. Those outcomes are then used, with
the fitted refinements unchanged, to study the resulting sharp intervals. A
separate analysis holds the unrepaired release fixed and examines realized AIPW
slack. HeartSteps provides a complementary example from an existing public
interface in which a pre-action summary is recorded alongside minute-level step
data, allowing us to ask whether temporal coarsening preserves a specified
threshold distinction. We begin with the release-level question: what must
survive for the specified evaluations to remain identified?

\section{Release exactness and worst-case ambiguity}
\label{sec:identification}

Let $H\in\cH$ denote the pre-action history,
$A\in\{1,\ldots,K\}$ the action, and $Y$ the observed outcome. A fixed
release map $R$ produces the released record $G=R(H)$, so the evaluator observes
$(G,A,Y)$. We call the joint law of these observed variables the
\emph{released law}. For each action $a$, let $Y(a)$ denote the potential
outcome under that action, with consistency $Y=Y(A)$. Histories mapped to one
released value $g$ form the \emph{release fiber}
\[
  \cH_R(g):=\{h\in\cH:R(h)=g\}.
\]

The behavior policy is $\mu(a\mid h)=\Pr(A=a\mid H=h)$ and may
vary across histories within a release fiber. We assume
$A\perp\!\!\!\perp\{Y(1),\ldots,Y(K)\}\mid H$ and
$\E|Y(a)|<\infty$. We fix the history space $\cH$, release map $R$,
behavior policy $\mu$, and target policies. An underlying law of
$(H,A,Y(1),\ldots,Y(K))$ is \emph{admissible} if its conditional action law
is $\mu(\cdot\mid H)$ and these assumptions hold.

\subsection{Evaluation weights and release exactness}

For a target policy $\pi(a\mid h)$, its value is
$V(\pi):=\E\{\sum_a\pi(a\mid H)Y(a)\}$. We consider target policies
$\pi_1,\ldots,\pi_M$ and a finite evaluation contract
$\mathcal E\subset\mathbb R^M$. 
Each $b\in\mathcal E$ specifies a requested value or comparison,
\[
  V_b
  :=
  \sum_{j=1}^M b_jV(\pi_j).
\]
For example, $b=e_j$ requests the value $V(\pi_j)$ and $b=e_j - e_k$ requests the contrast $V(\pi_j) - V(\pi_k)$, where $e_j$ denotes the $j$th standard basis vector of $\mathbb{R}^M$.

The release map $R$ is \emph{exact} for $\mathcal E$ if, for any released
law, all admissible underlying laws that induce it give the same $V_b$ for every
$b\in\mathcal E$.

For $b\in\mathcal E$, define the \emph{action-coefficient vector}
$\alpha_b(h)$ by
\[
  \alpha_b(h)_a
  :=
  \sum_j b_j\pi_j(a\mid h),
  \qquad
  V_b
  =
  \E\{\alpha_b(H)^\top Y(\cdot)\}.
\]
Thus its $a$th coordinate is the coefficient multiplying $Y(a)$ in $V_b$. Let $\mathcal A_{\mathcal E}$ denote the
actions with $\alpha_b(h)_a\ne0$ for some $b\in\mathcal E$ and $h\in\cH$, and assume
$\mu(a\mid h)>0$ for every $a\in\mathcal A_{\mathcal E}$ and $h\in\cH$.
For $b\in\mathcal E$ and $a\in\mathcal A_{\mathcal E}$, define the
\emph{evaluation weight}
\begin{equation*}
  w_b(h,a)
  :=
  \alpha_b(h)_a/\mu(a\mid h),
\end{equation*}
and set $w_b(h,a):=0$ on the remaining actions. Then
$V_b=\E\{w_b(H,A)Y\}$.

\begin{theorem}[Characterization of exact releases]
\label{thm:estimand-quotient}
The release map $R$ is exact for $\mathcal E$ if and only if, on every release
fiber $\cH_R(g)$, $w_b(h,a)$ is constant in $h$ for every
$b\in\mathcal E$ and $a\in\mathcal A_{\mathcal E}$.
Consequently, the map
\[
  h\longmapsto
  \bigl(w_b(h,a):b\in\mathcal E,\ a\in\mathcal A_{\mathcal E}\bigr)
\]
defines an exact release map. Two histories for which this collection differs
cannot share a record under any exact release. Thus the level sets of this map
give the coarsest exact release.
\end{theorem}

Exactness is evaluation-specific. The
release need not
reconstruct the entire pre-action history, only preserve distinctions that
change the evaluation weights $w_b(h,a)$. Two policy values may each be
ambiguous while their direct comparison is exact. A target policy can be
determined by $G$ while its value remains unidentified if those weights vary
within a release fiber.  Conversely, the behavior policy need not be a function
of $G$ when those weights are constant.

\subsection{Worst-case ambiguity}

Exactness is a binary property. It records whether the release can make a value or
comparison in the contract ambiguous, not how large that ambiguity can be. To quantify this
ambiguity within a release fiber, fix a released value $g$ and let
\[
  Q_g
  :=
  \operatorname{Law}(A,Y\mid G=g)
\]
denote the \emph{conditional released law} at $g$. A conditional released law
$Q_g$ is \emph{attainable} if it is induced by some admissible underlying law.
For $b\in\mathcal E$ and a fixed attainable $Q_g$, let
$[L_b(g;Q_g),U_b(g;Q_g)]$ be the sharp interval for the conditional evaluation
$\E\{\alpha_b(H)^\top Y(\cdot)\mid G=g\}$ over admissible underlying laws
inducing $Q_g$.

The width can vary across attainable $Q_g$, so we consider the largest possible
width.

\begin{theorem}[Worst-case ambiguity as an evaluation-weight radius]
\label{thm:general-evaluation-radius}
If $Y(a)\in[0,1]$ a.s., then for every released value $g$ and
$b\in\mathcal E$,
\[
  \sup_{Q_g}\{U_b(g;Q_g)-L_b(g;Q_g)\}
  =\Omega_g(b),
\]
where the supremum ranges over attainable $Q_g$ and
\begin{equation*}
  \Omega_g(b)
  :=
  \inf_{c\in\mathbb R^{|\mathcal A_{\mathcal E}|}}
  \sup_{h\in\cH_R(g)}
  \sum_{a\in\mathcal A_{\mathcal E}}
  \mu(a\mid h)\,|w_b(h,a)-c_a|.
\end{equation*}
\end{theorem}

The \emph{evaluation-weight radius} $\Omega_g(b)$ measures how well a single
action-indexed vector $c$ can approximate the history-specific evaluation
weights across the release fiber. It does not depend on any particular $Q_g$,
yet Theorem~\ref{thm:general-evaluation-radius} shows that it equals the
largest sharp width over attainable $Q_g$. This also recovers the exactness
criterion from Theorem~\ref{thm:estimand-quotient}. The release is exact for
$\mathcal E$ if and only if $\Omega_g(b)=0$ for every released value $g$ and
$b\in\mathcal E$. No compactness of $\cH_R(g)$ is required.

For any attainable $Q_g$, its sharp width is no larger than $\Omega_g(b)$ and
may be smaller. This distinction leads to two questions: how much the
worst-case ambiguity can be reduced by distinguishing histories more finely
within each released record, and how to characterize the sharp interval for a
particular $Q_g$.

\section{Finite-category release refinement}
\label{sec:adequacy}

To reduce the worst-case ambiguity, consider augmenting the release before evaluation
with one categorical label taking at most $m$ values within each release fiber.
We call the resulting release a \emph{finite-category refinement}.

\subsection{Optimal refinement}

For a released value $g$, let
$J_g:\cH_R(g)\to[m]$ be the label map for this refinement, so that the
refined record for $h\in\cH_R(g)$ is $(g,J_g(h))$. Equivalently, within the
fiber the same refinement is represented by the partition $\Pi_g$ into the
$r$ nonempty label classes of $J_g$, with $r\le m$:
\[
  \Pi_g=\{B_1,\ldots,B_r\},
  \qquad
  \cH_R(g)=\bigsqcup_{\ell=1}^r B_\ell.
\]

For a cell $B\in\Pi_g$, let $\Omega_B(b)$ denote the evaluation-weight radius in
Theorem~\ref{thm:general-evaluation-radius} with the release fiber restricted
to $B$. We minimize the largest $\Omega_B(b)$ over cells $B\in\Pi_g$ and directions
$b\in\mathcal E$:
\begin{equation}
  \chi_m(g;\mathcal E)
  :=
  \inf_{\Pi_g:\,|\Pi_g|\le m}
  \max_{B\in\Pi_g,\;b\in\mathcal E}
  \Omega_B(b).
  \label{eq:repair-frontier}
\end{equation}
When $m=1$ and $\mathcal E=\{b\}$, this reduces to
$\chi_1(g;\{b\})=\Omega_g(b)$. By
Theorem~\ref{thm:general-evaluation-radius}, $\chi_m(g;\mathcal E)$ is also
the smallest achievable worst-case sharp width across the refined cells and
directions in $\mathcal E$.

To characterize an optimal partition $\Pi_g$ attaining
$\chi_m(g;\mathcal E)$, recall that each radius $\Omega_B(b)$ minimizes over
a center $c_b\in\mathbb R^{\mathcal A_{\mathcal E}}$ for direction $b$.
Writing $c=(c_b)_{b\in\mathcal E}$, define the
\emph{pointwise distortion} $d(h,c)$ as the largest weighted deviation across
directions:
\[
  d(h,c)
  :=
  \max_{b\in\mathcal E}
  \sum_{a\in\mathcal A_{\mathcal E}}
  \mu(a\mid h)\,|w_b(h,a)-c_{b,a}|.
\]
Each $\Omega_B(b)$ minimizes over its own center $c_b$, so for $B\in\Pi_g$,
$\max_{b\in\mathcal E}\Omega_B(b)=\inf_c\sup_{h\in B}d(h,c)$.
Applying this identity across the cells of $\Pi_g$ yields an $m$-center problem
on $\cH_R(g)$.

\begin{theorem}[Optimal refinement as an $m$-center problem]
\label{thm:repair-frontier}
The optimal refinement value is
\[
  \chi_m(g;\mathcal E)
  =
  \min_{c_1,\ldots,c_m}
  \sup_{h\in\cH_R(g)}
  \min_{\ell\in[m]}d(h,c_\ell).
\]
For any minimizing centers $c_1^\star,\ldots,c_m^\star$, let
$\Pi_g^\star$ be the partition into at most $m$ cells induced by assigning each
$h\in\cH_R(g)$ to an index in
$\arg\min_{j\in[m]}d(h,c_j^\star)$, with ties broken arbitrarily. Then
$\Pi_g^\star$ is an optimal partition, attaining $\chi_m(g;\mathcal E)$.
\end{theorem}

\subsection{Finite-data repair and guarantees}

The finite-data repair uses histories
$H_1,\ldots,H_n\in\cH_R(g)$ to construct a refinement of the release fiber. For this construction, assume that the behavior policy is
constant on $\cH_R(g)$:
\[
  \mu(a\mid h)=\mu(a\mid h'),
  \quad
  a=1,\ldots,K,
  \quad
  h,h'\in\cH_R(g).
\]
Recall that $\alpha_b(h)$ is the action-coefficient vector, with
$\alpha_b(h)_a=\mu(a\mid h)w_b(h,a)$. Under this condition, the
$b$-specific term in $d(h,c)$ is an $\ell_1$ distance from $\alpha_b(h)$
to a center. Using another history $h'$ as the representative gives
$\|\alpha_b(h)-\alpha_b(h')\|_1$. We therefore use the distance
\[
  D(h,h')
  :=
  \max_{b\in\mathcal E}
  \|\alpha_b(h)-\alpha_b(h')\|_1.
\]

Restricting \eqref{eq:repair-frontier} to
$\{H_1,\ldots,H_n\}$ defines the \emph{finite-data optimum}
$\chi_{m,n}(g;\mathcal E)$, which serves as the benchmark for the partition
produced by Algorithm~\ref{alg:finite-data-repair}. The algorithm selects
representative histories by farthest-first traversal~\citep{gonzalez1985clustering}
using distance $D$, and evaluates the resulting candidate partitions under the
$d$-based objective.

\begin{algorithmbox}{Finite-data repair in a release fiber}
\label{alg:finite-data-repair}
\textbf{Input.} Histories $H_1,\ldots,H_n$; maximum categories $m$.

\smallskip
\textbf{Representatives.} For each initial representative $H_s$, $s\in[n]$,
initialize $S^{(s)}\leftarrow\{H_s\}$. While
$|S^{(s)}|<\min(m,n)$, choose $H^\star$, then update $S^{(s)}$:
\[
  H^\star
  \in
  \arg\max_{H_i\notin S^{(s)}}
  \min_{r\in S^{(s)}}D(H_i,r),
  \qquad
  S^{(s)}\leftarrow S^{(s)}\cup\{H^\star\}.
\]

\smallskip
\textbf{Candidate partitions.} For each $s\in[n]$, assign every $H_i$ to an
element of $\arg\min_{r\in S^{(s)}}D(H_i,r)$, with ties broken arbitrarily,
giving $\Pi^{(s)}$. Evaluate this partition by
\[
  C^{(s)}
  :=
  \max_{B\in\Pi^{(s)}}
  \min_c
  \max_{h\in B}d(h,c).
\]

\smallskip
\textbf{Selection.} Choose
$\widehat s\in\arg\min_{s\in[n]}C^{(s)}$, and set
$\widehat S:=S^{(\widehat s)}$ and
$\widehat\Pi:=\Pi^{(\widehat s)}$.

\smallskip
\textbf{Full-fiber extension.} For any $h\in\cH_R(g)$, assign $h$ to an
element of $\arg\min_{r\in\widehat S}D(h,r)$ using the same tie-breaking
rule. This extends the fitted refinement to $\cH_R(g)$.
\end{algorithmbox}

\begin{proposition}[Finite-data repair guarantee]
\label{prop:finite-data-repair}
For any $H_1,\ldots,H_n\in\cH_R(g)$,
Algorithm~\ref{alg:finite-data-repair} returns a partition $\widehat\Pi$
of these histories satisfying
\[
  \chi_{m,n}(g;\mathcal E)
  \le
  \max_{B\in\widehat\Pi,\;b\in\mathcal E}\Omega_B(b)
  \le
  2\chi_{m,n}(g;\mathcal E).
\]
Moreover, suppose $H_1,H_2,\ldots$ are i.i.d. draws from a distribution
on $\cH_R(g)$. If
\[
  \operatorname{supp}\bigl((\alpha_b(H_1))_{b\in\mathcal E}\bigr)
  =
  \overline{
    \bigl\{
      (\alpha_b(h))_{b\in\mathcal E}:h\in\cH_R(g)
    \bigr\}
  },
\]
then, for fixed $m$,
$\chi_{m,n}(g;\mathcal E)\to\chi_m(g;\mathcal E)$ almost surely.
\end{proposition}

Finite-data repair returns a partition with ambiguity at most twice the
finite-data optimum. Under further regularity, the bound converges to twice the
minimum worst-case ambiguity over full-fiber refinements.

Exact finite-data optimization is NP-hard.
Algorithm~\ref{alg:finite-data-repair} runs in polynomial time with a
factor-two guarantee, and no polynomial-time algorithm can guarantee a factor
strictly below two unless $\mathrm P=\mathrm{NP}$~\citep{feder1988clustering}.

The refinement criterion is outcome-free as it minimizes the largest ambiguity
over attainable $Q_g$. Once a particular $Q_g$ is fixed, the remaining
ambiguity can be smaller than this worst case.

\section{Sharp identification under fixed \texorpdfstring{$Q_g$}{Q\_g}}
\label{sec:transport}

Fix a released value $g$ and its conditional released law
$Q_g=\operatorname{Law}(A,Y\mid G=g)$. For an evaluation direction $b\in\mathcal E$, recall that
$[L_b(g;Q_g),U_b(g;Q_g)]$ is the sharp interval under $Q_g$. To characterize
this interval, we first determine which action-specific marginals are identified
by $Q_g$, and then characterize the remaining ambiguity.

For each action $a$, let
$F_a:=\operatorname{Law}\{Y(a)\mid G=g\}$ denote the distribution of the
potential outcome under action $a$. For every measurable outcome set $C$,
\begin{equation}
  Q_g(A=a,Y\in C)
  =
  \E\{\mu(a\mid H)\ind\{Y(a)\in C\}\mid G=g\}.
  \label{eq:fixed-law-weighted-row}
\end{equation}
When $\mu(a\mid h)$ varies over the fiber, different history-specific
distributions of $Y(a)$ can induce the same $Q_g$ while producing different
$F_a$.

\begin{proposition}[Identification of an action-specific marginal]
\label{prop:action-marginal-from-release}
For an action $a$ with $\mu(a\mid h)>0$ on $\cH_R(g)$, every attainable
$Q_g$ uniquely determines $F_a$ if and only if $\mu(a\mid h)$ is constant
over $h\in\cH_R(g)$.
\end{proposition}

When the constancy condition in
Proposition~\ref{prop:action-marginal-from-release} holds for every action and
release fiber, we call the behavior policy \emph{release-measurable}:
\begin{equation*}
  \mu(a\mid H)=\mu_a(G),
  \qquad a=1,\ldots,K.
\end{equation*}
At the fixed $g$, if $\mu_a(g)>0$ for every action,
Equation~\eqref{eq:fixed-law-weighted-row} becomes
$Q_g(A=a,Y\in C)=\mu_a(g)F_a(C)$, and hence
$F_a=\operatorname{Law}(Y\mid A=a,G=g)$ for $a=1,\ldots,K$.
If the behavior policy is not release-measurable at $g$, some $F_a$ need not
be determined by $Q_g$.  Appendix~\ref{app:general-behavior-sharp} gives the
corresponding bounded-outcome sharp formulation.

With $F_1,\ldots,F_K$ identified, the remaining task is to characterize
the sharp interval
$[L_b(g;Q_g),\allowbreak U_b(g;Q_g)]$, which is the range of
$\E\{\alpha_b(H)^\top Y(\cdot)\mid G=g\}$ over admissible underlying laws
inducing $Q_g$.

Conditional on $G=g$, $\alpha_b(H)$ takes values in
$\{\alpha_b(h):h\in\cH_R(g)\}$. For a given outcome vector $y$,
$\alpha_b(h)^\top y$ is linear in the coefficient vector, so its supremum
is unchanged if these values are replaced by their closed convex hull. We
thus define the \emph{action-coefficient set}
\begin{equation*}
  \mathcal C_b
  :=
  \overline{\operatorname{conv}}
  \{\alpha_b(h):h\in\cH_R(g)\}.
\end{equation*}
For this $y$, the fiberwise supremum is
\[
  \sigma_{\mathcal C_b}(y)
  :=
  \sup_{u\in\mathcal C_b}u^\top y
  =
  \sup_{h\in\cH_R(g)}\alpha_b(h)^\top y.
\]

For the potential-outcome vector $Y(\cdot)$, $Q_g$ identifies the marginals
$F_1,\ldots,F_K$, but not their joint law. Write
$\mathbf Y=(Y_1,\ldots,Y_K)\sim\gamma$, where
$\gamma\in\Gamma(F_1,\ldots,F_K)$, the class of all such couplings. Since
\[
  -\sigma_{\mathcal C_b}\{-Y(\cdot)\}
  \le
  \alpha_b(H)^\top Y(\cdot)
  \le
  \sigma_{\mathcal C_b}\{Y(\cdot)\},
\]
allowing all such outcome couplings gives outer bounds on the possible values of $\E\{\alpha_b(H)^\top Y(\cdot)\mid G=g\}$. We show that these bounds are sharp.

\begin{theorem}[Sharp interval over couplings of identified marginals]
\label{thm:master-transport}
Suppose the behavior policy is release-measurable, with $\mu_a(g)>0$ and
$\int |y|\,dF_a(y)<\infty$ for every action $a$. Then, for every
$b\in\mathcal E$,
\begin{equation}
\begin{aligned}
  U_b(g;Q_g)
  &=
  \sup_{\gamma\in\Gamma(F_1,\ldots,F_K)}
  \E_\gamma \sigma_{\mathcal C_b}(\mathbf Y),\\
  L_b(g;Q_g)
  &=
  -\sup_{\gamma\in\Gamma(F_1,\ldots,F_K)}
  \E_\gamma \sigma_{\mathcal C_b}(-\mathbf Y).
\end{aligned}
\label{eq:sharp-endpoints}
\end{equation}
\end{theorem}

Thus the histories $h\in\cH_R(g)$ matter for the sharp interval only through the set $\mathcal C_b$, and the remaining uncertainty is how
$F_1,\ldots,F_K$ are coupled. In particular, taking $b=e_j-e_k$ gives the
sharp interval for  $V(\pi_j)-V(\pi_k)$ directly. It is never wider than the interval derived from separate sharp intervals for the two
policy values.

Under release-measurable behavior, the action-specific constants
$\mu_a(g)$ rescale the evaluation weights into the action-coefficient vectors
defining $\mathcal C_b$, giving
\[
  \Omega_g(b)=\rho_1(\mathcal C_b),
\]
where $\rho_1$ denotes the $\ell_1$ Chebyshev radius. Thus $\mathcal C_b$
links the two ambiguity questions. Together with $F_1,\ldots,F_K$, it
determines the sharp interval for a fixed $Q_g$.  Its $\ell_1$ radius gives the
largest sharp width over attainable $Q_g$ on the normalized $[0,1]$ scale.
Consequently, on this scale,
$U_b(g;Q_g)-L_b(g;Q_g)\le\Omega_g(b)$.

\section{AIPW envelope and the sharp interval}
\label{sec:contract-geometry-bridge}

Section~\ref{sec:transport} characterizes the sharp interval for a fixed
$Q_g$ through the action-coefficient set $\mathcal C_b$, built from the
action-coefficient vectors $\alpha_b(h)$ within the release fiber. Under
release-measurable behavior, fix a coefficient vector $u\in\mathbb R^K$ and a
baseline vector $\xi=(\xi_1,\ldots,\xi_K)$, and let $e_a$ denote the $a$th
standard basis vector. The corresponding augmented inverse-probability-weighted
(AIPW) score is
\[
  u^\top\!\left(
    \xi+e_A\frac{Y-\xi_A}{\mu_A(g)}
  \right).
\]
Conditional on $H$ and $Y(\cdot)$, setting $u=\alpha_b(H)$ and averaging
this score over the behavior action recovers
$\alpha_b(H)^\top Y(\cdot)$ exactly. The coefficient $\alpha_b(H)$,
however, need not be the same across histories with $G=g$. We therefore bound
the score over $u\in\mathcal C_b$ to obtain the AIPW envelope and compare it
with the sharp interval.

\subsection{AIPW envelope and exact slack}

Under the conditions of Theorem~\ref{thm:master-transport}, fix a released
value $g$ and an evaluation direction $b\in\mathcal E$, and write
$\mu_a=\mu_a(g)$. For a baseline vector $\xi\in\mathbb R^K$, define
\[
  z_a(y;\xi)
  =
  \xi+e_a\frac{y-\xi_a}{\mu_a},
\]
so $u^\top z_A(Y;\xi)$ is the fixed-$u$ AIPW score displayed above.
Conditioning on $H$ and $Y(\cdot)$ gives
\[
  \E\!\left[
    \alpha_b(H)^\top z_A(Y;\xi)
    \mid H,Y(\cdot)
  \right]
  =
  \alpha_b(H)^\top Y(\cdot).
\]
This identity remains valid as $\alpha_b(H)$ varies within the release fiber.
Since its possible values lie in $\mathcal C_b$, the following bounds hold for
any baseline $\xi$:
\[
  -\sigma_{\mathcal C_b}\{-z_A(Y;\xi)\}
  \le
  \alpha_b(H)^\top z_A(Y;\xi)
  \le
  \sigma_{\mathcal C_b}\{z_A(Y;\xi)\}.
\]

Taking expectations given $G=g$ therefore bounds the conditional evaluation
$\E\{\alpha_b(H)^\top Y(\cdot)\mid G=g\}$. Under release-measurable
behavior, these bounds decompose into actionwise expectations under the
identified $F_a$. Optimizing over $\xi$ defines an \emph{AIPW envelope} with endpoints
\begin{equation}
\label{eq:aipw-envelope}
\begin{aligned}
  U_b^{\mathrm{AIPW}}(g)
  &=
  \inf_\xi
  \sum_a\mu_a
  \E_{F_a}\,\sigma_{\mathcal C_b}\{z_a(Y(a);\xi)\},\\
  L_b^{\mathrm{AIPW}}(g)
  &=
  -\inf_\xi
  \sum_a\mu_a
  \E_{F_a}\,\sigma_{\mathcal C_b}\{-z_a(Y(a);\xi)\}.
\end{aligned}
\end{equation}
The bounds hold for every admissible underlying law inducing $Q_g$, so the
AIPW envelope contains the sharp interval. The envelope applies the support
function action by action before averaging over the behavior action; computing
the envelope therefore requires no optimization over a joint coupling of the
$F_a$, unlike the sharp endpoints in Theorem~\ref{thm:master-transport}.

Applying the support function before this averaging can also create a gap. Since
$\sum_a\mu_a z_a(y_a;\xi)=y$, convexity gives the pointwise gap
\[
  S_b(\xi,y)
  :=
  \sum_a\mu_a
  \sigma_{\mathcal C_b}\{z_a(y_a;\xi)\}
  -
  \sigma_{\mathcal C_b}(y)
  \ge0.
\]
This gap vanishes exactly when a common $u\in\mathcal C_b$ maximizes
$u^\top z_a(y_a;\xi)$ for every action $a$.

We call the gap between corresponding AIPW and sharp endpoints
\emph{AIPW slack}. For the upper endpoint,
\begin{equation}
\label{eq:aipw-exact-gap}
  U_b^{\mathrm{AIPW}}(g)-U_b(g;Q_g)
  =
  \inf_{\xi,\;\gamma\in\Gamma(F_1,\ldots,F_K)}
  \E_\gamma S_b(\xi,\mathbf Y),
\end{equation}
and the lower endpoint follows by sign reversal. Thus the upper AIPW slack is
the infimum of the expected pointwise gap $S_b$ over the baseline $\xi$ and
outcome coupling $\gamma$. This slack measures the AIPW envelope's looseness relative to the sharp
interval under $Q_g$.

\subsection{Universal AIPW sharpness}

The exact characterization of AIPW slack in
\eqref{eq:aipw-exact-gap} makes the remaining dependence explicit. The slack
varies with the outcome marginals $F_1,\ldots,F_K$, while the relevant
variation over $h\in\cH_R(g)$ is already captured by $\mathcal C_b$. We next
characterize exactly when $\mathcal C_b$ makes the slack vanish for all outcome
marginals.

\begin{theorem}[Universal sharpness of the AIPW envelope]
\label{thm:aipw-universal-sharpness}
Fix $g$ and $b\in\mathcal E$. Suppose the behavior policy is
release-measurable, with $\mu_a(g)>0$ and
$\int |y|\,dF_a(y)<\infty$ for every action $a$. Then
\[
\begin{aligned}
  &U_b^{\mathrm{AIPW}}(g)=U_b(g;Q_g)
  \quad\text{for all }(F_1,\ldots,F_K)\\
  \Longleftrightarrow\;&
  L_b^{\mathrm{AIPW}}(g)=L_b(g;Q_g)
  \quad\text{for all }(F_1,\ldots,F_K)\\
  \Longleftrightarrow\;&
  \left|
    \{u_a:u\in\operatorname{ext}(\mathcal C_b)\}
  \right|
  \le2,
  \qquad a=1,\ldots,K.
\end{aligned}
\]
We call the last condition \emph{coordinatewise two-level}.
\end{theorem}

The AIPW envelope in Equation~\eqref{eq:aipw-envelope} evaluates the support
term $\sigma_{\mathcal C_b}\{z_a(Y(a);\xi)\}$ separately by action, so the
maximizing $u\in\mathcal C_b$ can vary with $a$ while all terms share the same
baseline $\xi$. Minimizing over $\xi$ leaves only a condition on the mean of
$u_a$ for each $a$; whether that mean determines how the values of $u_a$
across $\operatorname{ext}(\mathcal C_b)$ are combined depends on their number.

With two such values, the mean determines how mass is split between them. With
three or more, it need not: for $u_a\in\{0,1/2,1\}$, mean $1/2$ arises from
either all mass at $1/2$ or equal mass at $0$ and $1$. This extra actionwise
freedom can produce AIPW slack.

Theorem~\ref{thm:aipw-universal-sharpness} shows that when the two-level
condition fails, positive AIPW slack must occur under some admissible
underlying law, even though it can vanish under another. When each coordinate
is two-level, by contrast, the AIPW envelope in \eqref{eq:aipw-envelope}
always coincides with the sharp interval in \eqref{eq:sharp-endpoints}.

\section{Empirical consequences}
\label{sec:empirical}

This section examines the empirical consequences of the theoretical results in Sections~\ref{sec:identification}--\ref{sec:contract-geometry-bridge} across two complementary empirical settings using the KuaiRec and HeartSteps datasets.

KuaiRec's nearly complete response matrix \citep{gao2022kuairec} provides one setting in which the same OPE tasks are used across the analyses, with the behavior policy and pre-action release specified by construction.
We ask whether refinements matched in conditional entropy can differ in worst-case ambiguity, and whether those differences matter for sharp identification of pairwise policy comparisons once outcomes are observed.
We separately examine how the two-level condition relates to whether the AIPW envelope matches the sharp interval for the realized outcomes.

HeartSteps \citep{klasnja2019heartsteps} provides a temporal setting for asking whether coarsening minute-level step records preserves a threshold distinction that is resolved by a recorded pre-action summary.

\subsection{Worst-case ambiguity at matched conditional entropy}
\label{sec:empirical-kuairec}

Refinements with the same category limit $m$ need not yield
the same conditional entropy $H(\Pi\mid G)$. We therefore match refinements by this conditional entropy of
the refinement category given $G$, rather than by $m$ alone.

\begin{figure}[!t]
\centering
\includegraphics[width=0.70\columnwidth]{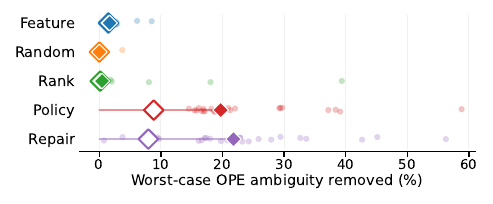}
\caption{\footnotesize \textbf{Matched conditional entropy does not equalize worst-case ambiguity.} Ambiguity is computed on the held-out evaluation group with outcomes withheld. For each task and direction, refined-cell ambiguity is averaged over the evaluation group and expressed as a percentage reduction from the unrepaired base; the task value is the smallest reduction across the ten contract directions. Small points, vertically jittered for visibility, show the 24 task values at the upper end of the conditional-entropy range common to every task--refinement pair. Open diamonds show medians across tasks halfway through this common range; filled diamonds show the medians at its upper end.}
\label{fig:kuairec-matched-information}
\end{figure}

Each of the 24 KuaiRec tasks defines a separate OPE instance. Its candidate
videos form the action set for $A$, complete play is the binary outcome $Y$,
and the behavior policy $\mu$ is uniform. Fine history $H$ is the pre-action
user-feature vector, while the base release $G$ retains two coarse activity
variables. The tasks vary in the number of candidate videos, their complete-play
rates among design users, and whether they come from one or several primary
content categories. Four target policies
(global, greedy, top-two, and softmax) define a ten-direction contract
consisting of their four values and six pairwise contrasts.
Appendix~\ref{app:kuairec-empirical} details task and policy construction.

Five refinement rules partition every release fiber of $G$ into at most $m$ categories: \emph{Feature}, pre-action
features; \emph{Random}, random partitions; \emph{Rank}, action-score ranks;
\emph{Policy-vector}, the target-policy vectors; and \emph{Repair},
Algorithm~\ref{alg:finite-data-repair} with the action-coefficient distance
from Section~\ref{sec:adequacy}.

Figure~\ref{fig:kuairec-matched-information} shows that at the largest common entropy,
the median ambiguity reduction is about $2\%$ or less for Feature,
Random, and Rank, versus about $20\%$ for Policy-vector and Repair. Thus
matching conditional entropy does not equalize worst-case ambiguity.

\subsection{Sharp intervals and sign certification}

We then use outcomes from the evaluation group, keeping the tasks, policies,
and fitted refinements fixed, to compute sharp intervals for the six pairwise
policy contrasts. This shifts the question from worst-case ambiguity to realized
sharp identification. A contrast is sign-certified when its sharp interval excludes
zero; the sharp intervals are identification intervals, not confidence
intervals.

\begin{figure}[!t]
\centering
\includegraphics[width=0.98\textwidth]{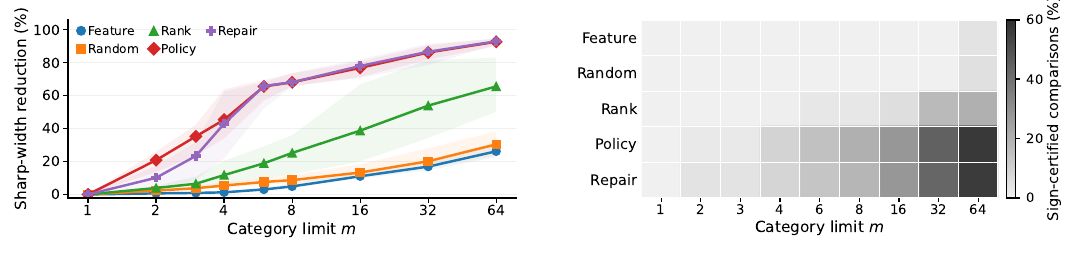}
\caption{\footnotesize \textbf{Narrower sharp intervals and more sign-certified policy comparisons.} Outcomes from the evaluation group are used while the tasks, policies, and fitted refinements remain fixed. \textbf{Left:} For each task and $m$, sharp-width reduction is based on the mean refined-to-unrepaired width ratio across the six policy contrasts; curves and bands show the median and interquartile range across 24 tasks. \textbf{Right:} Percentage of the 144 pairwise policy contrasts across 24 tasks whose sharp interval excludes zero.}
\label{fig:kuairec-consequence}
\end{figure}

In Figure~\ref{fig:kuairec-consequence}, Policy-vector and Repair yield about
$93\%$ median reduction in mean pairwise sharp width at $m=64$, versus about
$66\%$ for Rank and $26\%$--$30\%$ for Feature and Random. Sign certification
shows the corresponding policy-comparison consequence: $57\%$ at $m=64$ for
Policy-vector and Repair, versus $20\%$ for Rank and $4\%$--$5\%$ for Feature
and Random. At $m=8$, $18\%$--$20\%$ of comparisons are already sign-certified
for Policy-vector and Repair, versus none for Feature and Random. Every
certified sign agrees with the sign of the corresponding complete-data contrast
in the evaluation group.

\subsection{Two-level condition and realized AIPW slack}
\label{sec:empirical-aipw}

We now return to the unrepaired release $G$ and turn from sharp intervals to
AIPW slack. For each of the same 24 tasks and ten directions, we call a
task--direction pair two-level when the condition in
Theorem~\ref{thm:aipw-universal-sharpness} holds in every fiber of $G$, and
compare this classification with realized slack.

\medskip
{\centering
\footnotesize
\renewcommand{\arraystretch}{0.95}
\begin{tabular}{@{}lrr@{}}
\toprule
& Zero slack & Positive slack \\
\midrule
Two-level ($n=72$) & 72 & 0 \\
Multilevel ($n=168$) & 48 & 120 \\
\bottomrule
\end{tabular}\par}
\medskip

All 72 two-level pairs have zero slack, as guaranteed by
Theorem~\ref{thm:aipw-universal-sharpness}. Removing the 24 singleton
global-policy pairs leaves 48 non-singleton two-level pairs with median sharp
width about $0.994$, so theorem-guaranteed equality is not limited to zero-width
cases. Among the 168 multilevel pairs, 120 have positive slack and 48 have zero
slack; thus failure of the two-level condition does not imply positive realized
slack.

\subsection{Temporal coarsening and threshold preservation}
\label{sec:empirical-heartsteps}

HeartSteps V1 includes minute-level step counts and a recorded total for the
preceding 40 minutes \citep{klasnja2019heartsteps}; a later HeartSteps design
uses fewer than 150 steps in the preceding 40 minutes as a threshold
\citep{liao2018timing}. The analysis uses 6,590 time points with both records
available.

We remove the recorded 40-minute total, group the minute-level counts into
5-, 10-, 15-, or 30-minute bins, and shift the bin grid through every possible
one-minute offset. A time point is \emph{unresolved} if the grouped record does
not determine whether the 40-minute total is below the 150-step threshold.

\begin{table}[!ht]
\centering
\footnotesize
\setlength{\tabcolsep}{2.0pt}
\renewcommand{\arraystretch}{1.0}
\begin{tabular}{@{}lcccc@{}}
\toprule
Unresolved (\%) & 5 min & 10 min & 15 min & 30 min \\
\midrule
Offset range & 6.4--6.9 & 13.0--14.2 & 19.0--23.8 & 33.0--50.8 \\
Offset median & 6.6 & 13.4 & 19.8 & 48.9 \\
\bottomrule
\end{tabular}
\caption{\footnotesize \textbf{Unresolved time points after temporal coarsening.}
The recorded 40-minute total is removed. For each bin width, entries give the
range and median across one-minute grid offsets of the percentage of the 6,590
time points that are unresolved.}
\label{tab:heartsteps-ablation}
\end{table}

Table~\ref{tab:heartsteps-ablation} shows that even with 5-minute bins,
$6.4\%$--$6.9\%$ of time points remain unresolved across offsets. Yet adding back only the recorded 40-minute total resolves all time points. What matters is not simply how finely the history is recorded, but whether
the release preserves the distinction needed downstream.

With the original one-minute record, all 6,590 time points are resolved.
Even there, however, evaluating the same threshold at the exact recorded
timestamp changes whether the total is below 150 for $1.31\%$ of time points.

\section{Discussion}
\label{sec:discussion}

The release-level results turn the question of what must be preserved into a
concrete evaluation-specific structure. The action-specific evaluation weights
determine which histories can share a released record without losing
identification; on the normalized outcome scale, their behavior-weighted
within-record radius gives the largest sharp ambiguity the release can induce;
and the same radius becomes the finite-category refinement objective. Whether
identification is preserved, how severe its failure can be, and how refinement
should be targeted are therefore different consequences of the same
evaluation-specific distinctions. That loss is structural rather than a finite-sample limitation: additional
samples can better estimate the released law without recovering a
distinction the release no longer records.

The results also separate three layers that would otherwise be easy to collapse
into a single notion of uncertainty.
Worst-case ambiguity asks how much identification loss the release can induce across attainable released action--outcome laws. The sharp interval asks what remains once one such law is fixed. AIPW slack measures additional looseness of the envelope relative to that sharp benchmark, rather than additional release uncertainty.
This evaluation-specificity also appears in policy comparisons. At the release level,
cancellation can make a comparison exact even when its component values are
ambiguous. Under a fixed released law, evaluating the comparison directly keeps
its component values tied to the same underlying law and can yield a narrower
sharp interval than separate value bounds suggest. Release-measurable behavior is not a global premise. Exactness, worst-case
ambiguity, and ideal refinement allow behavior to vary within a released
record. The condition is used for the finite-data repair guarantee and, at
fixed law, to identify the action-specific marginals entering the sharp-interval
and AIPW analyses.

The empirical results give these distinctions concrete consequences. In
KuaiRec, refinements matched in conditional entropy could still differ in
worst-case OPE ambiguity. With those fitted refinements unchanged, using outcomes shifted the analysis
from worst-case release ambiguity to fixed-law sharp identification. Holding the
unrepaired release fixed further showed that failure of the
two-level condition does not by itself determine realized AIPW slack.
HeartSteps makes the preservation principle concrete in a different interface:
the recorded 40-minute aggregate preserves the specified threshold distinction
without retaining finer chronology, whereas temporal coarsening can lose it.
Across these examples, how much information a release retains---or how finely
it records time---does not by itself determine whether the distinctions needed
downstream are preserved.

The results also suggest a release-design perspective. Given an evaluation
contract and a prespecified tolerance for worst-case ambiguity, one can ask
which distinctions must be retained so that every requested value or comparison
has ambiguity below that tolerance. Section~\ref{sec:adequacy} makes this concrete for
finite-category refinements by characterizing the smallest achievable ambiguity
at each level of refinement. The same design problem could also account for
privacy by balancing how well the release preserves the distinctions needed for
the evaluation contract against how much sensitive information it reveals.

Taken together, the results answer the preservation question in
evaluation-specific terms. The release belongs to the identification problem
itself: downstream OPE can use only distinctions preserved by the release, yet exact support for the declared evaluations can require far less
than retaining the fine history.

\FloatBarrier

\bibliographystyle{unsrtnat}
\bibliography{tex/bib/references}

\clearpage
\appendix

\section{Release exactness and worst-case ambiguity}

This appendix proves the exact-release and worst-case ambiguity results from
Section~\ref{sec:identification} and records the corresponding scope
extensions. The behavior policy may vary within a release fiber unless a
subsection explicitly invokes the release-measurable specialization.

\begingroup
\let\SuppASubsection\subsection
\let\SuppASubsubsection\subsubsection
\renewcommand{\section}[1]{\SuppASubsection{Exact-release characterization}}
\let\subsection\SuppASubsubsection
\let\subsubsection\paragraph
\section{Exact-release characterization}
\label{app:identification-proofs}

Theorem~\ref{thm:estimand-quotient} follows from a fixed-distribution
criterion. We first fix one distribution of $(H,A,G)$ and then lift the
resulting almost-sure condition to a pointwise criterion on each release fiber.

\subsection{Fixed-distribution criterion}

For $b\in\mathcal E$, use the evaluation weight already defined in
Section~\ref{sec:identification},
\[
  w_b(h,a)=\frac{\alpha_b(h)_a}{\mu(a\mid h)}
\]
on contract-relevant actions, and set it to zero on the remaining actions. The
standing positivity condition gives
\[
  \E|w_b(H,A)|
  =\E\sum_a\left|\sum_j b_j\pi_j(a\mid H)\right|
  \le \sum_j|b_j|.
\]

\begin{lemma}[Fixed-distribution evaluation-weight criterion]
\label{lem:likelihood-quotient}
Under the OPE assumptions of Section~\ref{sec:identification}, fix
the distribution of $(H,A,G)$ and $b\in\mathcal E$.

\emph{(i)} If
\[
  w_b(H,A)=\E\{w_b(H,A)\mid G,A\}\quad\text{a.s.},
\]
then for any two admissible underlying laws $P$ and $P'$ sharing this
fixed distribution and having integrable potential outcomes,
\[
  \mathcal L_P(G,A,Y)=\mathcal L_{P'}(G,A,Y)
  \quad\Longrightarrow\quad
  V_b(P)=V_b(P').
\]

\emph{(ii)} If
\[
  \Pp\!\left[
    w_b(H,A)\ne\E\{w_b(H,A)\mid G,A\}
  \right]>0,
\]
then for every $0<\varepsilon<1/2$ there exist two admissible underlying
laws $P_+$ and $P_-$ sharing this fixed distribution, with binary
potential outcomes, such that
\[
  \left|
    \Pp_\pm(Y=1\mid H=h,A=a)-\frac12
  \right|\le\varepsilon
  \qquad\text{for all }(h,a),
\]
and
\[
  \mathcal L_{P_+}(G,A,Y)=\mathcal L_{P_-}(G,A,Y),
  \qquad
  V_b(P_+)\ne V_b(P_-).
\]
\end{lemma}

\paragraph{Proof.}
For part~(i), the inverse-probability identity and the assumption give
\[
  V_b
  =\E\!\left[
    Y\,\E\{w_b(H,A)\mid G,A\}
  \right].
\]
The conditional expectation in this display is determined by the fixed law of
$(H,A,G)$, so equality of the released laws of $(G,A,Y)$ makes this expectation
the same under $P$ and $P'$.

For part~(ii), define the proof-local centered residual and sign perturbation
\[
  \widetilde w_b=w_b-\E(w_b\mid G,A),
  \qquad
  \zeta_b=\operatorname{sgn}(\widetilde w_b)
  -\E\{\operatorname{sgn}(\widetilde w_b)\mid G,A\}.
\]
Then
\[
  |\zeta_b|\le2,
  \qquad
  \E(\zeta_b\mid G,A)=0,
  \qquad
  \E(\widetilde w_b\mid G,A)=0.
\]
Moreover,
\[
  \E(\zeta_bw_b)
  =\E(\zeta_b\widetilde w_b)
  =\E|\widetilde w_b|
   -\E\!\left[
      \E\{\operatorname{sgn}(\widetilde w_b)\mid G,A\}\widetilde w_b
    \right]
  =\E|\widetilde w_b|.
\]
The first equality uses $\E(\zeta_b\mid G,A)=0$; conditioning on $(G,A)$ in
the second term gives the last equality because
$\E(\widetilde w_b\mid G,A)=0$. By the assumption in part~(ii),
$\Pp(\widetilde w_b\ne0)>0$, so
$\E(\zeta_bw_b)=\E|\widetilde w_b|>0$.

Set
\[
  \theta_\pm(H,A)=\frac12\pm\frac{\varepsilon}{2}\zeta_b(H,A).
\]
Since $|\zeta_b|\le2$ and $0<\varepsilon<1/2$,
$\theta_\pm(H,A)\in[0,1]$ and
$|\theta_\pm(H,A)-1/2|\le\varepsilon$.
For each sign, take
$Y(a)\mid H=h\sim\operatorname{Bernoulli}\{\theta_\pm(h,a)\}$, with any
conditional coupling across actions, draw $A\mid H$ from the fixed behavior
policy independently of $Y(\cdot)$ conditional on $H$, and set $Y=Y(A)$. Then
\[
  \Pp_\pm(Y=1\mid G,A)
  =\E\{\Pp_\pm(Y=1\mid H,A)\mid G,A\}
  =\E\{\theta_\pm(H,A)\mid G,A\}
  =\frac12,
\]
where the last equality uses $\E(\zeta_b\mid G,A)=0$. Since the distribution
of $(G,A)$ is fixed and $Y$ is binary,
$\mathcal L_{P_+}(G,A,Y)=\mathcal L_{P_-}(G,A,Y)$.
Finally, the inverse-probability identity gives
$V_b(P_\pm)=\E\{\theta_\pm(H,A)w_b(H,A)\}$, so
\[
  V_b(P_+)-V_b(P_-)
  =\E\!\left[\{\theta_+(H,A)-\theta_-(H,A)\}w_b(H,A)\right]
  =\varepsilon\,\E(\zeta_bw_b)>0.
\]
\qed

Only histories with positive probability under the fixed distribution enter the
almost-sure condition in Lemma~\ref{lem:likelihood-quotient}.

\subsection{From a fixed distribution to the fiberwise criterion}

\begin{lemma}[Fiberwise evaluation-weight criterion]
\label{lem:fiberwise-weight-criterion}
Suppose that, for each nonempty release fiber $\cH_R(g)$, every probability
distribution of $H$ supported on that fiber is admissible. Then the release
identifies $V_b$ for every attainable released law if and only if, for every
$g$, every action $a$ used by the direction, and every
$h,h'\in\cH_R(g)$,
\[
  w_b(h,a)=w_b(h',a).
\]
\end{lemma}

\paragraph{Proof.}
If the weights are constant on each relevant fiber--action pair, then
$w_b(H,A)$ is a function of $(G,A)$ under every admissible underlying law,
so Lemma~\ref{lem:likelihood-quotient}(i) gives identification. Conversely, if
the weights differ at two histories in one fiber for a used action, choose a
history distribution assigning positive probability to both histories. Then
$\Pp[w_b(H,A)\ne\E\{w_b(H,A)\mid G,A\}]>0$, and
Lemma~\ref{lem:likelihood-quotient}(ii) gives the nonidentification witness.
\qed

The fixed-distribution condition is almost sure under the chosen history law;
the release-wide condition must hold at every history in each admissible
release fiber.

\paragraph{Proof of Theorem~\ref{thm:estimand-quotient}.}
If the vector
$\bigl(w_b(h,a):b\in\mathcal E,\ a\in\mathcal A_{\mathcal E}\bigr)$
is constant on each release fiber, then
Lemma~\ref{lem:fiberwise-weight-criterion} identifies $V_b$ for every
$b\in\mathcal E$.

Conversely, if two histories in one release fiber give different values of
this vector, then some $b\in\mathcal E$ and
contract-relevant action $a$ satisfy
$w_b(h,a)\ne w_b(h',a)$. Lemma~\ref{lem:fiberwise-weight-criterion} then
shows that $V_b$ is not identified from the release. Therefore every exact
pre-action release separates histories with different values of this vector,
while its level-set partition is itself exact. \qed

The theorem therefore also identifies the coarsest exact release.

\paragraph{Span invariance.}
The evaluation contract may contain linearly redundant directions. For this
observation only, extend $\alpha_b$ and $w_b$ to
$b\in\operatorname{span}(\mathcal E)$ by the same linear formulas. Because
$b\mapsto w_b$ is linear, such redundancies do not affect the coarsest exact
release.

\begin{corollary}[Span invariance of the exact release]
\label{cor:exact-span-invariance}
Let $\beta_1,\ldots,\beta_d$ be any basis of
$\operatorname{span}(\mathcal E)$. For any histories $h,h'$,
\[
  w_b(h,a)=w_b(h',a)
  \ \forall b\in\mathcal E,\ a\in\mathcal A_{\mathcal E}
  \quad\Longleftrightarrow\quad
  w_{\beta_\ell}(h,a)=w_{\beta_\ell}(h',a)
  \ \forall \ell,\ a\in\mathcal A_{\mathcal E}.
\]
Hence $\mathcal E$ and any basis of its span induce the same coarsest exact
release.
\end{corollary}

\paragraph{Proof.}
Both implications follow from linearity in $b$. The final claim follows from
Theorem~\ref{thm:estimand-quotient}. \qed

\endgroup

\begingroup
\let\SuppASubsection\subsection
\let\SuppASubsubsection\subsubsection
\renewcommand{\section}[1]{\SuppASubsection{Worst-case ambiguity characterization}}
\let\subsection\SuppASubsubsection
\let\subsubsection\paragraph
\section{Worst-case ambiguity characterization}
\label{app:general-behavior-radius}

This section proves Theorem~\ref{thm:general-evaluation-radius} by maximizing
the sharp width over attainable $Q_g$. The radius $\Omega_g(b)$ is normalized
to an outcome interval of length one.

Fix one released value $g$ and one direction $b\in\mathcal E$. On
contract-relevant actions,
\[
  w_b(h,a)=\frac{\alpha_b(h)_a}{\mu(a\mid h)}.
\]
For $c\in\mathbb R^K$, write only for this proof
\[
  d_a(h;c)=\alpha_b(h)_a-\mu(a\mid h)c_a.
\]
Then the radius is equivalently
\begin{equation*}
  \Omega_g(b)
  =\inf_{c\in\mathbb R^K}
    \sup_{h\in\cH_R(g)}
    \sum_{a\in\mathcal A_{\mathcal E}}|d_a(h;c)|.
\end{equation*}
For $a\notin\mathcal A_{\mathcal E}$, every $b\in\mathcal E$ has
$\alpha_b(h)_a=0$, and those coordinates may be taken with $c_a=0$. Hence the unqualified action sums in
the finite-LP calculation below may run over all $K$ actions without changing
the radius.

For outcomes in $[0,B]$, we prove
\begin{equation}
  \sup_{Q_g}
  \{U_b(g;Q_g)-L_b(g;Q_g)\}
  =B\,\Omega_g(b),
  \label{eq:general-radius-identity}
\end{equation}
where the supremum is over attainable $Q_g$.

\subsection{Finite-fiber identity}

By scaling, take $B=1$. It is enough to consider potential outcomes in
$\{0,1\}$. Indeed, take any two admissible underlying laws $P,Q$ with outcomes in
$[0,1]$ that induce the same $Q_g$, and under each law define
\[
  X(a)\mid H,Y(a)\sim\operatorname{Bernoulli}\{Y(a)\},
\]
using auxiliary randomization independent of $A$ conditional on $H,Y(\cdot)$,
and set $X=X(A)$. Then
\[
  \E\{X(a)\mid H,Y(a)\}=Y(a),
\]
so every target expectation is preserved. Moreover,
\[
  \Pr(A=a,X=1\mid G=g)
  =\E\{\mu(a\mid H)Y(a)\mid G=g\}
  =\E\{\ind(A=a)Y\mid G=g\},
\]
which is determined by their common $Q_g$; the action masses then determine
the corresponding probabilities at $X=0$. Thus the two transformed laws induce
the same conditional law of $(A,X)\mid G=g$ while preserving their target
separation. Every separation attainable with $[0,1]$ outcomes is therefore
attainable with endpoint-binary outcomes; the reverse inclusion is immediate.
Hence, for the remainder of the proof, we restrict to endpoint-binary
potential outcomes.

For a fixed $Q_g$, its sharp width is the largest value separation between
two admissible underlying laws inducing that same $Q_g$. Maximizing this
separation over attainable $Q_g$ is therefore equivalent to maximizing over
admissible law pairs that induce a common released law. Suppose first that
$\cH_R(g)$ is finite; this is a finite linear program.
With dual variables $\varphi_a(0),\varphi_a(1)$ and two normalization constants,
write
\[
  u_a=\varphi_a(0),
  \qquad
  c_a=\varphi_a(1)-\varphi_a(0),
\]
and define
\[
  P_h(c)=\sum_a[d_a(h;c)]_+,
  \qquad
  N_h(c)=\sum_a[-d_a(h;c)]_+.
\]
After optimizing the normalization constants, the dual value is
\begin{equation}
  \max_h\left\{P_h(c)-\sum_a\mu(a\mid h)u_a\right\}
  +\max_h\left\{N_h(c)+\sum_a\mu(a\mid h)u_a\right\}.
  \label{eq:general-radius-dual-intermediate}
\end{equation}
For every $h$,
\[
  P_h(c)+N_h(c)=\sum_a|d_a(h;c)|,
\]
so \eqref{eq:general-radius-dual-intermediate} is at least
$\max_h\sum_a|d_a(h;c)|$. Let
\[
  s_b=\sum_{a=1}^K\alpha_b(h)_a=\sum_jb_j,
\]
which does not depend on $h$. Then
\[
\begin{aligned}
  P_h(c)&=\frac12\left\{\sum_a|d_a(h;c)|+s_b
      -\sum_a\mu(a\mid h)c_a\right\},\\
  N_h(c)&=\frac12\left\{\sum_a|d_a(h;c)|-s_b
      +\sum_a\mu(a\mid h)c_a\right\}.
\end{aligned}
\]
At $u=-c/2$, \eqref{eq:general-radius-dual-intermediate} equals
\[
  \frac12\max_h\left\{\sum_a|d_a(h;c)|+s_b\right\}
  +\frac12\max_h\left\{\sum_a|d_a(h;c)|-s_b\right\}
  =\max_h\sum_a|d_a(h;c)|.
\]
Finite LP strong duality therefore gives
\[
  \sup_{Q_g}\{U_b(g;Q_g)-L_b(g;Q_g)\}
  =\inf_c\max_h\sum_a|d_a(h;c)|
  =\Omega_g(b)
\]
for a finite release fiber. The next subsection removes the finiteness
assumption and then records the zero-radius and release-measurable
specializations.

\subsection{Arbitrary-fiber extension}
\label{app:general-behavior-radius-compact}

The remaining step extends the finite-fiber identity to arbitrary release
fibers. For finite $F\subset\cH_R(g)$, define
\[
  \Omega_{g,F}(b)
  =\inf_c\max_{h\in F}
    \sum_{a\in\mathcal A_{\mathcal E}}
      |\alpha_b(h)_a-\mu(a\mid h)c_a|.
\]
By the preceding finite-fiber argument, $\Omega_{g,F}(b)$ is the largest
sharp width attainable by laws supported on $F$.

First,
\begin{equation}
  \Omega_g(b)
  =\sup_{F\subset\cH_R(g):\,F\text{ finite}}
    \Omega_{g,F}(b).
  \label{eq:finite-subset-radius-limit}
\end{equation}
The left side is at least the right side because restricting the history set
cannot increase the center problem. For the reverse inequality, fix
$h_\star\in\cH_R(g)$ and suppose some $t<\Omega_g(b)$ satisfies
$\Omega_{g,F}(b)\le t$ for every finite $F$ containing $h_\star$. Define
\[
  C_h(t)
  =\left\{c:
    \sum_{a\in\mathcal A_{\mathcal E}}
      |\alpha_b(h)_a-\mu(a\mid h)c_a|\le t
    \right\}
\]
on the contract-relevant coordinates. Positivity makes $C_{h_\star}(t)$
compact. By the assumption on $\Omega_{g,F}(b)$, every finite family
\[
  \{C_h(t)\cap C_{h_\star}(t):h\in F\}
\]
has nonempty intersection, so compactness gives
\[
  \bigcap_{h\in\cH_R(g)}C_h(t)\ne\varnothing.
\]
This would produce a full-fiber center of radius at most $t$, contradicting
$t<\Omega_g(b)$. Hence \eqref{eq:finite-subset-radius-limit} holds.

For any $\varepsilon>0$, choose finite $F$ with
\[
  \Omega_{g,F}(b)>\Omega_g(b)-\varepsilon.
\]
The finite-fiber result gives two underlying endpoint-binary laws supported on
$F$ that induce the same $Q_g$ and whose values differ by
$\Omega_{g,F}(b)$, so the worst-case width over the full fiber is at least
$\Omega_g(b)-\varepsilon$.

For the reverse bound, take any endpoint-binary pair inducing the same $Q_g$ and
any center $c$. For contract-relevant actions put
\[
  \varphi_a(0)=-\frac{c_a}{2},
  \qquad
  \varphi_a(1)=\frac{c_a}{2},
\]
and set $\varphi_a\equiv0$ on the remaining actions. Because the two laws induce
the same $Q_g$, the expectations of every function of the observed pair $(A,Y)$
agree. For $x\in\{0,1\}^K$,
\[
  \alpha_b(h)^\top x
  -\sum_a\mu(a\mid h)\varphi_a(x_a)
\]
lies in
\[
  \left[
    \frac{s_b-D_c(h)}{2},
    \frac{s_b+D_c(h)}{2}
  \right],
\]
where
\[
  s_b=\sum_jb_j,
  \qquad
  D_c(h)
  =\sum_{a\in\mathcal A_{\mathcal E}}
    |\alpha_b(h)_a-\mu(a\mid h)c_a|.
\]
Because the two laws have the same $Q_g$, the $\varphi$-term has the same
expectation under both, so the interval above bounds their conditional-value
difference by $\sup_hD_c(h)$. Taking the infimum over $c$ gives the upper
bound $\Omega_g(b)$. Letting
$\varepsilon\downarrow0$ proves \eqref{eq:general-radius-identity} for
outcomes in $[0,1]$, and scaling gives the factor $B$. \qed

\paragraph{Zero radius.}
The center problem attains its infimum on the contract-relevant coordinates:
fixing $h_\star\in\cH_R(g)$, pointwise positivity makes the objective at
$h_\star$ coercive in $c$. Hence $\Omega_g(b)=0$ if and only if one vector
$c$ satisfies
\[
  w_b(h,a)=c_a
  \qquad
  \text{for every }h\in\cH_R(g)
  \text{ and contract-relevant action }a,
\]
which is exactly the fiberwise condition in
Theorem~\ref{thm:estimand-quotient}.

\subsection{Release-measurable behavior}

If $\mu(a\mid h)=\mu_a(g)>0$ throughout the release fiber, then the coordinate
change $\eta_a=\mu_a(g)c_a$ gives
\[
  \Omega_g(b)
  =\inf_c\sup_h\sum_a|\alpha_b(h)_a-\mu_a(g)c_a|
  =\inf_{\eta\in\mathbb R^K}\sup_h\|\alpha_b(h)-\eta\|_1.
\]
Convexification and closure do not change the supremum of
$u\mapsto\|u-\eta\|_1$, hence
\[
  \Omega_g(b)=\rho_1(\mathcal C_b).
\]
This is the release-measurable specialization of the radius.

\endgroup

\section{Finite-category refinement and finite-data repair}

This appendix proves the finite-category refinement and finite-data repair
results from Section~\ref{sec:adequacy}. The ideal refinement problem allows
the behavior policy to vary within a release fiber. Within-fiber constant
behavior is required only for the action-coefficient specialization used by the
finite-data repair procedure and its guarantees.

\subsection{Attainment without a compact history space}
\label{app:finite-label-b6}

We first justify the attainment statement in
Theorem~\ref{thm:repair-frontier} without assuming that the release fiber is
compact. Fix one release fiber $\cH_R(g)$ and the finite evaluation
contract $\mathcal E$. For a finite set $F\subset\cH_R(g)$, let
$\chi_{m,F}(g;\mathcal E)$ denote the $m$-category objective from
Section~\ref{sec:adequacy} with the history fiber restricted to $F$. To keep
the proof notation compact, write the main-text
center collection $c=(c_b)_{b\in\mathcal E}$ as $C$ and set
$d_{g,\mathcal E}(h,C):=d(h,c)$. Define the sublevel set
\[
  K_h(t)=\{C:d_{g,\mathcal E}(h,C)\le t\},
  \qquad t\ge0.
\]
Each $K_h(t)$ is nonempty and compact. For finite $F$,
\[
  \chi_{m,F}(g;\mathcal E)\le t
\]
if and only if there are at most $m$ centers such that, for every $h\in F$,
at least one center lies in $K_h(t)$.

\paragraph{Finite-to-full extension.}
Suppose the preceding condition holds for every finite subfamily of an indexed
family $\{K_i:i\in I\}$ of compact sets. For each finite subfamily, assign
every set to one of the at most $m$ centers it contains. These finite-coloring
constraints have the finite-intersection property in the compact product space
$[m]^I$, yielding a global coloring for which every finite same-color
subfamily has nonempty intersection. Compactness then gives one center in the
intersection of each nonempty color class.

\begin{lemma}[Finite-subset reduction and refinement attainment]
\label{lem:finite-label-attainment}
Under the assumptions of Theorem~\ref{thm:repair-frontier},
\[
  \chi_m(g;\mathcal E)
  =\sup_{F\subset\cH_R(g):\,F\text{ finite}}
    \chi_{m,F}(g;\mathcal E),
\]
and the fiberwise minimum in Theorem~\ref{thm:repair-frontier} is attained.
Consequently, for every $\varepsilon>0$, there is a finite
$F\subset\cH_R(g)$ such that
\[
  \chi_{m,F}(g;\mathcal E)
  >\chi_m(g;\mathcal E)-\varepsilon.
\]
\end{lemma}

\paragraph{Proof.}
Restricting the history fiber cannot increase the objective, so
\[
  \sup_{F\text{ finite}}\chi_{m,F}(g;\mathcal E)
  \le\chi_m(g;\mathcal E).
\]
Let the left side be $t_\star$. For every finite $F$, at most $m$ centers can
be chosen so that every $K_h(t_\star)$, $h\in F$, contains one of them. The
finite-to-full extension above therefore gives at most $m$ centers with the
same property over the full release fiber. Assign each history to one such center
in $K_h(t_\star)$. The resulting partition has worst radius at most
$t_\star$, hence
\[
  \chi_m(g;\mathcal E)\le t_\star.
\]
This proves equality and attainment. The final finite-witness statement follows
from the definition of the supremum. \qed

Under ordinary measurability of the known behavior and target-policy
coordinates, indexing the finitely many attaining centers and breaking ties by
the smallest index gives a measurable cell assignment within the fiber. The
result does not assert a globally measurable selection of optimal centers over
an arbitrary uncountable release space.

\begingroup
\let\SuppBSubsection\subsection
\let\SuppBSubsubsection\subsubsection
\renewcommand{\section}[1]{\SuppBSubsection{Finite-center characterization and finite-data repair}}
\let\subsection\SuppBSubsubsection
\let\subsubsection\paragraph
\section{Finite-center characterization and finite-data repair}
\label{app:contract-quantization}

With attainment established, we first prove
Theorem~\ref{thm:repair-frontier} by identifying the cost of one partition
cell and then passing between partitions and collections of at most $m$
centers. We then specialize the characterization to within-fiber constant
behavior; the following subsubsections use that specialization for the
finite-data procedure, its guarantees, and transfer to the full refinement
domain. Continue the notation from Appendix~\ref{app:finite-label-b6}:
$C=(c_b)_{b\in\mathcal E}$ and
$d_{g,\mathcal E}(h,C)=d(h,c)$. The finite-center argument itself allows the
known behavior policy to vary within the release fiber.

\paragraph{Proof of Theorem~\ref{thm:repair-frontier}.}
For a nonempty partition cell $B$, write only for this proof
\[
  f_b(h,c_b)
  =\sum_{a\in\mathcal A_{\mathcal E}}
    \mu(a\mid h)|w_b(h,a)-c_{b,a}|.
\]
By definition,
\[
  d_{g,\mathcal E}(h,C)=\max_{b\in\mathcal E}f_b(h,c_b).
\]
Because each evaluation direction has its own center $c_b$, the centers can
be optimized jointly as $C$; since $\mathcal E$ is finite, the maximum over
directions and supremum over histories may be interchanged:
\[
  \max_{b\in\mathcal E}\inf_{c_b}\sup_{h\in B}f_b(h,c_b)
  =\inf_{C=(c_b)}\max_{b\in\mathcal E}\sup_{h\in B}f_b(h,c_b)
  =\inf_C\sup_{h\in B}d_{g,\mathcal E}(h,C).
\]
Thus the cost of one partition cell is exactly its best one-center distortion.

Given a partition into at most $m$ nonempty cells, choose one optimal center
for each cell. Conversely, given $C_1,\ldots,C_m$, assign each history to an
index attaining
\[
  \min_{\ell\le m}d_{g,\mathcal E}(h,C_\ell),
\]
breaking ties by the smallest index. Using these centers already gives the
displayed pointwise cost, and reoptimizing within the induced cells can only
decrease it. Therefore
\[
 \chi_m(g;\mathcal E)
 =
 \inf_{C_1,\ldots,C_m}
 \sup_{h\in\cH_R(g)}
 \min_{\ell\le m}d_{g,\mathcal E}(h,C_\ell).
\]
Lemma~\ref{lem:finite-label-attainment} shows that the infimum is attained,
which gives the minimum stated in Theorem~\ref{thm:repair-frontier}. \qed

\paragraph{Within-fiber constant behavior.}
Now suppose
\[
  \mu(a\mid h)=\mu_a(g)
  \qquad(h\in\cH_R(g)).
\]
By the standing positivity condition, $\mu_a(g)>0$ for
$a\in\mathcal A_{\mathcal E}$. On these active coordinates, the change
\[
  \eta_{\ell,b,a}=\mu_a(g)c_{\ell,b,a}
\]
is bijective; set the inactive coordinates of $\eta_{\ell,b}$ to zero. Then
\[
  d_{g,\mathcal E}(h,C_\ell)
  =\max_{b\in\mathcal E}
    \|\alpha_b(h)-\eta_{\ell,b}\|_1.
\]
Thus, under this specialization, Theorem~\ref{thm:repair-frontier} is the
$m$-center problem for the stacked vectors
$(\alpha_b(h))_{b\in\mathcal E}$ under the product metric
\[
  \max_{b\in\mathcal E}\|\alpha_b(h)-\alpha_b(h')\|_1.
\]

The same representation gives a category-count interpretation: the smallest
$m$ for which $\chi_m(g;\mathcal E)\le\varepsilon$ is the minimum number
of radius-$\varepsilon$ balls in this metric needed to cover the stacked
vectors $(\alpha_b(h))_{b\in\mathcal E}$. At $\varepsilon=0$, this is the
number of distinct stacked vectors represented in the fiber.

\subsection{Finite-data repair procedure}
\label{app:practical-contract-repair}

Under within-fiber constant behavior and the standing positivity condition on
$\mathcal A_{\mathcal E}$, Appendix~\ref{app:contract-quantization} reduces
the finite-data instance of Section~\ref{sec:adequacy} to an $m$-center
problem. Given design histories $H_1,\ldots,H_n$ in one base-release fiber,
duplicate coefficient vectors do not affect the maximum-radius objective, so
let $h_1,\ldots,h_q$ represent their distinct stacked vectors
$(\alpha_b(h))_{b\in\mathcal E}$. Define their pairwise distance by
\[
  d(i,j)=\max_{b\in\mathcal E}
    \|\alpha_b(h_i)-\alpha_b(h_j)\|_1.
\]
For these represented design histories, the finite-data optimum is exactly
\[
  \chi_{m,n}(g;\mathcal E)
  =\min_{\eta_1,\ldots,\eta_m}
    \max_{i\le q}\min_{\ell\le m}
      \max_{b\in\mathcal E}\|\alpha_b(h_i)-\eta_{\ell,b}\|_1,
\]
where $\eta_\ell=(\eta_{\ell,b})_{b\in\mathcal E}$ is one ambient center.

\paragraph{Procedure.}
Let $m'=\min(m,q)$. From every represented starting point $s$, run
farthest-first traversal in the metric $d$ \citep{gonzalez1985clustering}
until $m'$ representatives have been selected. Assign each represented history
to its nearest selected representative. For each resulting cell $I$, recompute
its best ambient radius
\[
  \min_{(\eta_b)_{b\in\mathcal E}}
    \max_{i\in I}\max_{b\in\mathcal E}
      \|\alpha_b(h_i)-\eta_b\|_1,
\]
which is a finite linear program. Score each starting point by the maximum of
these cell radii and retain the lowest-cost candidate. The score is the exact
Section~\ref{sec:adequacy} objective of that candidate partition.

An implementation may additionally score prefixes with fewer than $m'$
representatives and carry forward a better lower-category candidate. Doing so
can only improve the fitted objective and is not needed for the factor-two
guarantee in Proposition~\ref{prop:finite-data-repair}. No direction-wise
rescaling is used.

\subsection{Finite-data guarantees and computational boundary}
\label{app:finite-repair-guarantees}

This subsection proves Proposition~\ref{prop:finite-data-repair} and its
instance-specific fitting bounds, then records the computational boundary of
the represented finite-data problem.

Fix one base-release value $g$ and assume the behavior policy is constant
within the fiber, with the standing positivity condition on
$\mathcal A_{\mathcal E}$.
Given design histories $H_1,\ldots,H_n$ in this fiber, write
\[
  x(h)=\bigl(\alpha_b(h)\bigr)_{b\in\mathcal E},
  \qquad
  d\{x(h),x(h')\}
  =
  \max_{b\in\mathcal E}
  \|\alpha_b(h)-\alpha_b(h')\|_1 ,
\]
and let $S_n=\{x_1,\ldots,x_q\}$ be the distinct vectors represented among
$x(H_1),\ldots,x(H_n)$. For a partition $\Pi$ of $S_n$ into at most $m$ cells, let
$J_{S_n}(\Pi)$ be its exact Section~\ref{sec:adequacy} refinement cost, with
ambient cell centers allowed. The finite-data optimum is then
\[
  \chi_{m,n}(g;\mathcal E)
  =
  \min_{\Pi:\,|\Pi|\le m}J_{S_n}(\Pi).
\]
Thus $\chi_{m,n}$ is the oracle optimum for the histories available at design
time; the fitted repair partition need not attain it.

For $m<q$ and a starting point $s\in S_n$, let
$c_1^{(s)},\ldots,c_{m+1}^{(s)}$ be the first $m+1$ points selected by
farthest-first traversal, and define the radius after the first $m$
representatives by
\[
  r_m(s)
  =
  \max_{x\in S_n}\min_{\ell\le m}d\{x,c_\ell^{(s)}\}.
\]
For $m<q$, set
\[
  L_m=\frac12\max_{s\in S_n}r_m(s),
  \qquad
  R_m=\min_{s\in S_n}r_m(s).
\]
For $m\ge q$, set $L_m=R_m=U_m=0$ and retain all $q$ represented
points. For $m<q$, let $U_m$ be the smallest exact refinement cost, across all
starting points, after nearest-representative assignment with $m$
representatives and exact cellwise recentering.

For the finite-data repair procedure in Section~\ref{sec:adequacy}, retain
the main-text notation $\widehat S$ for the selected representatives and
$\widehat\Pi$ for the selected partition, and write
$U_m=J_{S_n}(\widehat\Pi)$.

\begin{proposition}[Finite-data optimality bracket]
\label{prop:finite-repair-bracket}
For every fixed design sample and every $m\ge1$,
\[
  L_m
  \le
  \chi_{m,n}(g;\mathcal E)
  \le
  J_{S_n}(\widehat\Pi)
  =
  U_m
  \le
  R_m
  \le
  2L_m.
\]
Hence
\[
  \chi_{m,n}(g;\mathcal E)
  \le
  J_{S_n}(\widehat\Pi)
  \le
  2\chi_{m,n}(g;\mathcal E),
\]
which is the first claim of Proposition~\ref{prop:finite-data-repair}.
\end{proposition}

\paragraph{Proof.}
The case $m\ge q$ is immediate. Suppose $m<q$ and fix a start $s$.
After the first $m$ representatives have been selected, let $c_{m+1}^{(s)}$ be a point
farthest from them. Farthest-first cover radii are nonincreasing, so
\[
  d\{c_i^{(s)},c_j^{(s)}\}\ge r_m(s)
  \qquad
  (1\le i<j\le m+1).
\]
Thus these $m+1$ represented points are pairwise separated by at least
$r_m(s)$. If $m$ ambient balls of radius strictly below $r_m(s)/2$ covered
$S_n$, two of these $m+1$ points would lie in the same ball, and the triangle
inequality would make their distance strictly smaller than $r_m(s)$.
Therefore
\[
  \chi_{m,n}(g;\mathcal E)\ge \frac{r_m(s)}2
\]
for every start $s$, which gives $L_m\le\chi_{m,n}(g;\mathcal E)$.

Now take a start attaining $R_m$. Assign every represented point to a nearest
selected representative. Using that representative itself as a feasible center
gives every cell radius at most $R_m$. Recomputing the best ambient center in
each fixed cell cannot increase the refinement cost. Because the procedure selects
the lowest exact recentered cost across all starts, $U_m\le R_m$. The selected
partition uses at most $m$ cells, so
$\chi_{m,n}(g;\mathcal E)\le U_m$. Finally
\[
  R_m
  \le
  \max_s r_m(s)
  =
  2L_m.
\]
\qed

The lower and upper bounds in Proposition~\ref{prop:finite-repair-bracket} are
computed during fitting; the guarantee does not require solving the exact
$m$-center problem.

\paragraph{Computation.}
If $B=|\mathcal E|$ and $K$ is the number of actions, the pairwise metric on
$q$ distinct represented vectors costs $O(q^2BK)$ arithmetic operations and
$O(q^2)$ memory. If refinements are computed for all budgets up to a maximum
$M$, then, given this matrix, running the farthest-first path from every
represented start through the largest effective budget $\min(M,q)$ costs
$O\{q^2\min(M,q)\}$. For a fixed cell containing $n_I$ represented points, the exact
$\ell_1$ one-center problem for one evaluation direction is a linear program
with $K+n_IK+1$ variables and $O(n_IK)$ inequalities; the directions separate.
Thus the finite-data repair procedure runs in polynomial time in the
represented input size under the standard linear-programming model.

We next compare this polynomial-time guarantee with the complexity of exact
finite-data refinement.

\begin{proposition}[Hardness of exact finite-data refinement]
\label{prop:finite-repair-hardness}
Even with one release fiber, two actions, a contract containing two policy
values, uniform behavior, and a finite history space, exact finite-data
refinement is NP-hard.
Moreover, unless $\mathrm P=\mathrm{NP}$, no polynomial-time algorithm can
guarantee an approximation factor strictly below two over this class.
\end{proposition}

\paragraph{Proof.}
Reduce from central $m$-center in the planar $\ell_\infty$ metric, for which
Feder and Greene~\citep{feder1988clustering} give the corresponding
factor-two hardness. After a common translation and positive rescaling, take
the input points to be $p_i=(p_{i1},p_{i2})\in[0,1]^2$.
Create one history $h_i$ per point, use two actions with uniform behavior,
and define two target policies by
\[
  \pi_j(1\mid h_i)=p_{ij},
  \qquad
  \pi_j(2\mid h_i)=1-p_{ij},
  \qquad j=1,2.
\]
Let the contract contain the two policy values. Then
\[
  \alpha_{e_j}(h_i)=(p_{ij},1-p_{ij}).
\]
For any cell $I$ and direction $j$, the smallest ambient $\ell_1$ radius of
these coefficient vectors is
\[
  \max_{i\in I}p_{ij}-\min_{i\in I}p_{ij}.
\]
Indeed, the $\ell_1$ diameter is twice this range, giving the lower bound, and
the midpoint on the line $\{(p,1-p):p\in\mathbb R\}$ attains it. Hence the
refinement radius of $I$ is
\[
  \max_{j=1,2}
  \left\{
    \max_{i\in I}p_{ij}-\min_{i\in I}p_{ij}
  \right\},
\]
which is exactly twice the continuous $\ell_\infty$ one-center radius of the
planar points in $I$. Minimizing over at most $m$ cells therefore preserves
approximation ratios. The planar $\ell_\infty$ hardness transfers directly.
\qed

The farthest-first backbone is classical~\citep{gonzalez1985clustering};
here the OPE release problem determines the metric and cell objective to which
it is applied.

\subsection{Transfer to the full refinement domain}
\label{app:finite-repair-transfer}

The preceding guarantees concern optimization over the represented set
$S_n$. We now separate this finite-data problem from how well $S_n$ covers
the full refinement domain. Let
\[
  \mathcal X_g
  =
  \overline{
    \left\{
      x(h):h\in\mathcal H_R(g)
    \right\}
  }
\]
be the closure of the attainable stacked action-coefficient vectors in the
fiber. Because the center objective is continuous in these vectors, taking the
closure does not change the full-fiber optimum, so
$\chi_m(\mathcal X_g)=\chi_m(g;\mathcal E)$. The observed set
$S_n\subseteq\mathcal X_g$ has directed coverage error
\[
  \Delta(S_n,\mathcal X_g)
  =
  \sup_{x\in\mathcal X_g}\min_{s\in S_n}d(x,s).
\]

\begin{proposition}[Deterministic repair stability]
\label{prop:finite-repair-stability}
For every $m$,
\[
  0
  \le
  \chi_m(g;\mathcal E)-\chi_{m,n}(g;\mathcal E)
  \le
  \Delta(S_n,\mathcal X_g).
\]
Consequently,
\[
  L_m
  \le
  \chi_m(g;\mathcal E)
  \le
  U_m+\Delta(S_n,\mathcal X_g).
\]
\end{proposition}

\paragraph{Proof.}
Because $S_n\subseteq\mathcal X_g$, restricting the point set cannot increase
the $m$-center optimum, so
$\chi_{m,n}(g;\mathcal E)\le\chi_m(g;\mathcal E)$. Let $m$ ambient
centers cover $S_n$ with radius $\chi_{m,n}(g;\mathcal E)$. For any
$x\in\mathcal X_g$, choose $s\in S_n$ within
$\Delta(S_n,\mathcal X_g)$ of $x$. The triangle inequality gives a center
within
$\chi_{m,n}(g;\mathcal E)+\Delta(S_n,\mathcal X_g)$ of $x$. Taking the
supremum over $x$ proves the upper bound. Proposition~\ref{prop:finite-repair-bracket} gives the final
display. \qed

Proposition~\ref{prop:finite-repair-stability} compares the finite- and
full-domain optimal values. To control the fitted partition itself under the
nearest-representative extension, use the retained representative set
$\widehat S$. Every $x\in\mathcal X_g$ is within
\[
  \max_{s\in S_n}\min_{r\in\widehat S}d(s,r)
  +\Delta(S_n,\mathcal X_g)
\]
of its nearest retained representative. Using those representatives as
feasible cell centers gives
\begin{equation*}
  J_{\mathcal X_g}(\widehat\Pi_{g,m})
  \le
  \max_{s\in S_n}\min_{r\in\widehat S}d(s,r)
  +\Delta(S_n,\mathcal X_g).
\end{equation*}
This is the transfer bound used by the empirical implementation. The
representative-coverage term above need not equal the smaller exact recentered
cost $U_m$, especially when a lower-budget partition is carried forward.

\paragraph{Fixed-fiber sampling specialization.}
The preceding results are deterministic. A sampling statement requires an
additional relation between the design distribution and the intended repair
domain. For this paragraph only, hold the release map, behavior rule, target
policies, and evaluation contract fixed, and suppose that $H_1,H_2,\ldots$
are i.i.d. from a design distribution on this fiber. Let their stacked vectors
have distribution $\nu_g$ with support $\mathcal X_g$. Let
$S_n$ be the represented sample and
\[
  \Delta_n
  =
  \Delta(S_n,\mathcal X_g).
\]
For $r>0$, let $N_g(r)$ be the radius-$r$ covering number of
$\mathcal X_g$ and
\[
  p_g(r)
  =
  \inf_{x\in\mathcal X_g}
  \nu_g\{B_d(x,r)\}.
\]
The coefficient vectors are bounded, so $\mathcal X_g$ is compact; full
support then implies $p_g(r)>0$ for every fixed $r>0$. A union bound over a
radius-$\varepsilon/2$ cover gives
\begin{equation*}
  \Pr(\Delta_n>\varepsilon)
  \le
  N_g(\varepsilon/2)
  \exp\{-n\,p_g(\varepsilon/2)\}.
\end{equation*}
Therefore $\Delta_n\to0$ almost surely and
\[
  \chi_{m,n}(g;\mathcal E)\longrightarrow\chi_m(g;\mathcal E)
  \qquad\text{a.s.}
\]
For the actual finite-data repair procedure, on the event
$\Delta_n\le\varepsilon$,
Propositions~\ref{prop:finite-repair-bracket}--\ref{prop:finite-repair-stability}
give
\[
  L_m
  \le
  \chi_m(g;\mathcal E)
  \le
  U_m+\varepsilon,
\]
while the fitted nearest-representative rule satisfies
\[
  J_{\mathcal X_g}(\widehat\Pi_{g,m})
  \le
  \max_{s\in S_n}\min_{r\in\widehat S}d(s,r)+\varepsilon.
\]
This is a fixed-fiber sampling specialization. The release-wide theory does
not assume that every admissible history lies in the support of a design
distribution, nor that repeated i.i.d. histories are available within every
released value $g$.

\endgroup

\section{Sharp identification and AIPW analysis}

This appendix first gives the sharp formulation for a fixed $Q_g$ when
behavior varies within a release fiber. It then proves the coupling
characterization in Theorem~\ref{thm:master-transport} and the AIPW results
from Section~\ref{sec:contract-geometry-bridge}. These coupling and AIPW
results use release-measurable behavior and integrable action-specific outcome
marginals. In the convex-analytic proofs, a generic set
$\mathcal S=\mathcal C_b$ and its support functions keep the notation compact;
$\mathcal C_b$ remains the corresponding OPE object.

\begingroup
\let\SuppCSubsection\subsection
\let\SuppCSubsubsection\subsubsection
\renewcommand{\section}[1]{\SuppCSubsection{Sharp formulation with within-fiber behavior variation}}
\let\subsection\SuppCSubsubsection
\let\subsubsection\paragraph
\section{Sharp formulation with within-fiber behavior variation}
\label{app:general-behavior-sharp}

This subsection records the fixed-$Q_g$ sharp problem before imposing
release-measurable behavior. Fix $G=g$, an evaluation direction $b$, and the
conditional released law
\[
  Q_g=\operatorname{Law}(A,Y\mid G=g).
\]
Consider a candidate conditional law
$\tau(dh,dy_1,\ldots,dy_K)$ for
$(H,Y(1),\ldots,Y(K))\mid G=g$, supported on $\cH_R(g)$ in the history
coordinate. Compatibility with $Q_g$ requires, for every action $a$ and
measurable outcome set $C$, the weighted row constraints
\begin{equation}
  \int \mu(a\mid h)\ind\{y_a\in C\}\,d\tau
  =\Pr(A=a,Y\in C\mid G=g).
  \label{eq:weighted-row-constraint}
\end{equation}
The right-hand side is the observed action-$a$ row of $Q_g$;
Equation~\eqref{eq:weighted-row-constraint} introduces no additional observed
object. The general sharp problem is therefore optimization over candidate
laws satisfying these compatibility constraints.

\begin{proposition}[Sharp interval with within-fiber behavior variation]
\label{prop:general-behavior-sharp}
Under the Section~\ref{sec:identification} conditions and bounded potential
outcomes, the sharp endpoints satisfy
\begin{equation}
  U_b(g;Q_g)=\sup_{\tau}\int \alpha_b(h)^\top y\,d\tau,
  \qquad
  L_b(g;Q_g)=\inf_{\tau}\int \alpha_b(h)^\top y\,d\tau.
  \label{eq:general-behavior-primal}
\end{equation}
where the supremum and infimum range over conditional laws $\tau$ in the
stated bounded-outcome model class, supported on $\cH_R(g)$ and satisfying
Equation~\eqref{eq:weighted-row-constraint}.
\end{proposition}

\paragraph{Proof.}
Take any admissible underlying law inducing $Q_g$ and let $\tau$ be the
conditional law of $(H,Y(1),\ldots,Y(K))$ given $G=g$. By the definition of $\alpha_b$,
its conditional evaluation is
\[
  \E\!\left\{\alpha_b(H)^\top Y(\cdot)\mid G=g\right\}
  =\sum_{a=1}^K\int \alpha_b(h)_a y_a\,d\tau
  =\int \alpha_b(h)^\top y\,d\tau.
\]
The same conditional law $\tau$ must also reproduce the observed action--outcome rows.
Indeed, consistency and conditional independence give, for every $a$ and $C$,
\[
\begin{aligned}
  \Pr(A=a,Y\in C\mid G=g)
  &=\E\!\left[
    \ind\{A=a\}\ind\{Y(a)\in C\}\mid G=g
  \right]\\
  &=\E\!\left[
    \mu(a\mid H)\ind\{Y(a)\in C\}\mid G=g
  \right]\\
  &=\int\mu(a\mid h)\ind\{y_a\in C\}\,d\tau.
\end{aligned}
\]
Thus every admissible underlying law inducing $Q_g$ yields a feasible $\tau$
whose conditional evaluation equals the objective in
Equation~\eqref{eq:general-behavior-primal}.

Conversely, take any feasible $\tau$. Draw
$(H,Y(1),\ldots,Y(K))\mid G=g$ from $\tau$, draw $A\mid H$ from the known
behavior policy independently of $Y(\cdot)$ conditional on $H$, and set
$Y=Y(A)$. The support condition ensures $R(H)=g$, the construction satisfies
the stated OPE assumptions, and Equation~\eqref{eq:weighted-row-constraint}
makes the induced law of $(A,Y)\mid G=g$ exactly $Q_g$. Its conditional
evaluation is the displayed objective. Hence the feasible objective values are
exactly the conditional evaluations attainable under $Q_g$, so the two
extrema are sharp. \qed

The same weighted constraints also characterize when an action-specific
outcome marginal is itself identified.

\paragraph{Proof of Proposition~\ref{prop:action-marginal-from-release}.}
If $\mu(a\mid h)=\mu_a(g)>0$ throughout the fiber, then for every measurable
outcome set $C$,
\[
  \Pr(A=a,Y\in C\mid G=g)
  =\mu_a(g)F_a(C),
\]
so the action-$a$ row of $Q_g$ determines $F_a$.

Conversely, suppose two histories $h_1,h_2$ in the fiber satisfy
$m_1:=\mu(a\mid h_1)\ne m_2:=\mu(a\mid h_2)$. Give $h_1$ and $h_2$ equal
conditional probability given $G=g$ and keep this history law fixed across the
two admissible underlying laws. Set every potential outcome other than $Y(a)$ to zero. Under the first law let $Y(a)$ be Bernoulli$(1/2)$ at both histories. Under the second, for
sufficiently small $\delta>0$, let its success probabilities at
$(h_1,h_2)$ be
\[
  \frac12+\delta m_2,
  \qquad
  \frac12-\delta m_1.
\]
The perturbations are chosen to leave the behavior-weighted success mass
unchanged:
\[
  \frac12m_1\!\left(\frac12+\delta m_2\right)
  +\frac12m_2\!\left(\frac12-\delta m_1\right)
  =\frac{m_1+m_2}{4},
\]
the same value as under the first law. The action masses and all other
action--outcome rows are also unchanged, so the two laws induce the same
$Q_g$. Their unweighted $Y(a)$ marginals differ, however,
because the second law's marginal success probability is
\[
  \frac12+\frac{\delta}{2}(m_2-m_1)\ne\frac12.
\]
Hence $F_a$ is not determined by $Q_g$. \qed

\paragraph{Release-measurable behavior.}
If the condition in Proposition~\ref{prop:action-marginal-from-release} holds for
every action, write
\[
  \mu(a\mid h)=\mu_a(g)>0
  \qquad(h\in\cH_R(g)).
\]
Then every action-specific outcome marginal $F_a$ is identified, and
Equation~\eqref{eq:weighted-row-constraint} reduces to the marginal
constraints
\[
  \tau\{Y(a)\in C\}=F_a(C).
\]
Thus the observed law constrains the outcome coordinates of $\tau$ through
$F_1,\ldots,F_K$. Separately, for a fixed potential-outcome vector $y$, the
history-side optimization is the pointwise identity
\[
  \sup_{h\in\cH_R(g)}\alpha_b(h)^\top y
  =\sigma_{\mathcal C_b}(y).
\]
This equality is pointwise in $y$ and does not require a maximizing history to
exist. These two reductions lead to the coupling formulation in
Theorem~\ref{thm:master-transport}; Lemma~\ref{lem:approx-support-realization}
gives the finite-net realization used to establish sharpness without requiring
a measurable maximizing history.

\endgroup

\begingroup
\let\SuppCSubsection\subsection
\let\SuppCSubsubsection\subsubsection
\renewcommand{\section}[1]{\SuppCSubsection{Sharp coupling characterization}}
\let\subsection\SuppCSubsubsection
\let\subsubsection\paragraph
\section{Sharp coupling characterization}
\label{app:transport-proofs}

Under release-measurable behavior, Appendix~\ref{app:general-behavior-sharp}
reduces compatibility with $Q_g$ to the identified marginals
$F_1,\ldots,F_K$. We now prove the resulting sharp coupling
characterization. Its only auxiliary step is a finite-net realization that
avoids assuming a measurable maximizing history. We then record the implication
for direct policy comparisons.

\paragraph{Proof of Theorem~\ref{thm:master-transport}.}
Fix $G=g$ and an evaluation direction $b$, and write
$\mathcal S=\mathcal C_b$ for the action-coefficient set. For compact
notation let
$h_{\mathcal S}(y)=\sigma_{\mathcal S}(y)$ and
$\ell_{\mathcal S}(y)=-\sigma_{\mathcal S}(-y)$.
For $h\in\cH_R(g)$ and a potential-outcome vector $y$,
\[
  \alpha_b(h)^\top y
  \le h_{\mathcal S}(y).
\]
Hence every admissible underlying law inducing $Q_g$, with
induced coupling $\gamma\in\Gamma(F_1,\ldots,F_K)$, satisfies
\[
  \E\{\alpha_b(H)^\top Y(\cdot)\mid G=g\}
  \le
  \E_\gamma h_{\mathcal S}(\mathbf Y)
  \le
  \sup_{\gamma'\in\Gamma(F_1,\ldots,F_K)}
  \E_{\gamma'}h_{\mathcal S}(\mathbf Y).
\]

For the converse, fix
$\gamma\in\Gamma(F_1,\ldots,F_K)$. Under release-measurable behavior,
compatibility with $Q_g$ fixes the action-specific marginals but imposes no
additional restriction on their association with a history in $\cH_R(g)$.
Lemma~\ref{lem:approx-support-realization} therefore gives, for every
$\delta>0$, an admissible underlying law inducing $Q_g$ whose potential-outcome
coupling is $\gamma$ and whose conditional evaluation is at least
\[
  \E_\gamma h_{\mathcal S}(\mathbf Y)
  -\delta\E_\gamma\|\mathbf Y\|_\infty.
\]
The first-moment assumption implies
$\E_\gamma\|\mathbf Y\|_\infty<\infty$, so letting $\delta\downarrow0$ and
then taking the supremum over $\gamma$ proves the formula for
$U_b(g;Q_g)$. Applying the same argument to the lower support function gives
\[
  L_b(g;Q_g)
  =\inf_{\gamma\in\Gamma(F_1,\ldots,F_K)}
    \E_\gamma\ell_{\mathcal S}(\mathbf Y)
  =
  -\sup_{\gamma\in\Gamma(F_1,\ldots,F_K)}
    \E_\gamma h_{\mathcal S}(-\mathbf Y).
\]
\qed

\par
The approximation step used in the converse is isolated in the following lemma.

\begin{lemma}[Approximate realization of the support bound]
\label{lem:approx-support-realization}
Fix $G=g$, a direction $b$, and
$\gamma\in\Gamma(F_1,\ldots,F_K)$. Under release-measurable behavior, for
every $\delta>0$ there exists an admissible underlying law inducing $Q_g$,
with potential-outcome coupling $\gamma$, whose conditional evaluation is at
least
\[
  \E_\gamma \sigma_{\mathcal C_b}(\mathbf Y)
  -\delta\E_\gamma\|\mathbf Y\|_\infty.
\]
\end{lemma}

\paragraph{Proof.} The attainable action-coefficient vectors
\[
  \{\alpha_b(h):h\in\cH_R(g)\}
\]
are bounded because $\|\alpha_b(h)\|_1\le\sum_j|b_j|$. For this $\delta$, choose actual histories $h_1,\ldots,h_N$ whose
coefficient vectors form a $\delta$-net of this set in $\ell_1$. Then
\[
  \sigma_{\mathcal C_b}(y)
  \le \max_i \alpha_b(h_i)^\top y+\delta\|y\|_\infty.
\]
Choose the smallest maximizing index $i(y)$. Because the candidate set is
finite, $i(y)$ is measurable. Draw $\mathbf Y\sim\gamma$, set
$H=h_{i(\mathbf Y)}$, draw the action from the release-measurable behavior
policy, and set the observed outcome to $Y=Y(A)$. Each selected history lies
in $\cH_R(g)$, and for every action $a$ and measurable set $C$,
\[
  \Pr(A=a,Y\in C\mid G=g)
  =\mu_a(g)F_a(C),
\]
so the constructed law induces the same $Q_g$. Its conditional evaluation
satisfies
\[
  \E\{\alpha_b(H)^\top Y(\cdot)\mid G=g\}
  \ge
  \E_\gamma \sigma_{\mathcal C_b}(\mathbf Y)
  -\delta\E_\gamma\|\mathbf Y\|_\infty.
\]
Thus neither compactness of the release fiber nor a measurable maximizing
history is required. Convexification and closure add no assumption because
they leave the support function unchanged. \qed

\paragraph{Direct policy comparisons.}
Theorem~\ref{thm:master-transport} applies to the requested direction itself,
so a policy contrast is optimized under one common underlying law rather than
by combining separate policy-value extrema. For $b=e_j-e_k$, let
$[L_j,U_j]$ and $[L_k,U_k]$ denote the two sharp value intervals under the
same $Q_g$. Then
\[
  [L_{e_j-e_k}(g;Q_g),U_{e_j-e_k}(g;Q_g)]
  \subseteq
  [L_j-U_k,\,U_j-L_k].
\]
The inclusion can be strict because the extrema defining the separate value
intervals need not be jointly attainable under one underlying law. This is an
identification statement, not a sampling-variance comparison.

\endgroup

\begingroup
\let\SuppCSubsection\subsection
\let\SuppCSubsubsection\subsubsection
\renewcommand{\section}[1]{\SuppCSubsection{AIPW envelope and universal sharpness}}
\let\subsection\SuppCSubsubsection
\let\subsubsection\paragraph
\section{AIPW envelope and universal sharpness}
\label{app:aipw-core-proofs}

Throughout this subsection, use a generic set $\mathcal S$, instantiated
by $\mathcal C_b$, and write
$h_{\mathcal S}(y)=\sigma_{\mathcal S}(y)$. Let
$L_{\mathcal S},U_{\mathcal S}$ denote the sharp endpoints
$L_b(g;Q_g),U_b(g;Q_g)$, and let
$A^-_{\mathcal S},A^+_{\mathcal S}$ denote
$L_b^{\mathrm{AIPW}}(g),U_b^{\mathrm{AIPW}}(g)$. We first prove the
outer-bound and exact-slack relations. We then prove universal sharpness:
Lemma~\ref{lem:aipw-mean-coherence} isolates the finite-support AIPW
constraint, after which sufficiency and necessity are treated separately.

\subsection{AIPW outer bound and exact slack}

Fix a nonempty compact set $\mathcal S$. For a baseline
$\xi\in\mathbb R^K$, recall
\[
  z_a(y;\xi)=\xi+e_a\frac{y-\xi_a}{\mu_a},
\]
so that, for every outcome vector $y$,
\[
  \sum_a\mu_a z_a(y_a;\xi)=y.
\]
Convexity of the support function gives
\[
  \sum_a\mu_a h_{\mathcal S}\{z_a(y_a;\xi)\}
  \ge h_{\mathcal S}(y).
\]
Taking expectations and optimizing over $\xi$ gives
$U_{\mathcal S}\le A^+_{\mathcal S}$; sign reversal gives
$A^-_{\mathcal S}\le L_{\mathcal S}$. Hence
\[
  A^-_{\mathcal S}\le L_{\mathcal S}
  \le U_{\mathcal S}\le A^+_{\mathcal S}.
\]

For a fixed baseline, write $A^+_{\mathcal S}(\xi)$ for the unoptimized upper
objective in Equation~\eqref{eq:aipw-envelope}, and define
\[
  J_{\mathcal S}(\xi,y)
  =
  \sum_a\mu_a h_{\mathcal S}\{z_a(y_a;\xi)\}
  -h_{\mathcal S}(y)
  \ge0.
\]
For any $\gamma\in\Gamma(F_1,\ldots,F_K)$,
\[
  A^+_{\mathcal S}(\xi)
  -\E_\gamma h_{\mathcal S}(\mathbf Y)
  =
  \E_\gamma J_{\mathcal S}(\xi,\mathbf Y).
\]
Using the sharp coupling formula from Theorem~\ref{thm:master-transport},
\[
\begin{aligned}
  A^+_{\mathcal S}-U_{\mathcal S}
  &=
  \inf_\xi A^+_{\mathcal S}(\xi)
    -\sup_{\gamma\in\Gamma(F_1,\ldots,F_K)}
       \E_\gamma h_{\mathcal S}(\mathbf Y)\\
  &=
  \inf_\xi\inf_{\gamma\in\Gamma(F_1,\ldots,F_K)}
    \E_\gamma J_{\mathcal S}(\xi,\mathbf Y).
\end{aligned}
\]
This is Equation~\eqref{eq:aipw-exact-gap}.

\paragraph{Pointwise equality.}
Using $\sum_a\mu_a z_a(y_a;\xi)=y$,
\[
\begin{aligned}
  J_{\mathcal S}(\xi,y)
  &=
  \sum_a\mu_a h_{\mathcal S}\{z_a(y_a;\xi)\}
  -
  \max_{u\in\mathcal S}
  u^\top\sum_a\mu_a z_a(y_a;\xi)\\
  &=
  \min_{u\in\mathcal S}
  \sum_a\mu_a
  \left[
    h_{\mathcal S}\{z_a(y_a;\xi)\}
    -u^\top z_a(y_a;\xi)
  \right].
\end{aligned}
\]
Every summand is nonnegative. Since $\mathcal S$ is compact and every
$\mu_a>0$, $J_{\mathcal S}(\xi,y)=0$ if and only if one
$u\in\mathcal S$ maximizes $u^\top z_a(y_a;\xi)$ for every action $a$.
The lower-endpoint counterpart follows by sign reversal. In particular,
\[
  A^+_{\mathcal S}=U_{\mathcal S}
  \quad\Longleftrightarrow\quad
  \inf_{\xi,\gamma}
  \E_\gamma J_{\mathcal S}(\xi,\mathbf Y)=0.
\]

\subsection{Proof of Theorem~\ref{thm:aipw-universal-sharpness}}

The proof of Theorem~\ref{thm:aipw-universal-sharpness} starts from a
finite-support reformulation of the AIPW envelope. Under the two-level
condition, its rowwise selectors can be represented by one common selector;
when the condition fails, two geometric lemmas produce a finite-support
positive-slack witness. We first isolate the finite-support AIPW constraint.

\begin{lemma}[AIPW row-mean primal]
\label{lem:aipw-mean-coherence}
Let $\mathcal S$ be nonempty, compact, and convex, let
$\mu=(\mu_1,\ldots,\mu_K)$ be a strictly positive probability vector, and
suppose
\[
  F_a=\sum_{i=1}^{n_a}p_{ai}\delta_{y_{ai}}.
\]
For $u_{ai}\in\mathcal S$, write
\[
  s^{(a)}=\sum_i p_{ai}u_{ai},
  \qquad
  \bar s=\sum_{a=1}^K\mu_a s^{(a)}.
\]
Then
\[
  A_{\mathcal S}^{+}
  =
  \max
  \left\{
    \sum_{a=1}^K\sum_{i=1}^{n_a}
      p_{ai}(u_{ai})_a y_{ai}:
    u_{ai}\in\mathcal S,
    \ s^{(a)}_a=\bar s_a\ \text{for every }a
  \right\}.
\]
\end{lemma}

\paragraph{Proof.}
For fixed $\xi$,
\[
  A_{\mathcal S}^{+}(\xi)
  =
  \max_{u_{ai}\in\mathcal S}
  \Bigg\{
    \sum_{a,i}p_{ai}(u_{ai})_a y_{ai}
    +\xi^\top
      \left[
        \sum_a\mu_a s^{(a)}
        -(s^{(1)}_1,\ldots,s^{(K)}_K)^\top
      \right]
  \Bigg\}.
\]
The product of the copies of $\mathcal S$ is compact and convex, and the
objective is affine in the selectors and $\xi$. Sion's minimax theorem permits
interchanging $\inf_\xi$ and the selector maximum. The infimum over
$\xi\in\mathbb R^K$ is finite exactly when
\[
  \sum_a\mu_a s^{(a)}
  =(s^{(1)}_1,\ldots,s^{(K)}_K)^\top,
\]
which is equivalent to
\[
  s^{(a)}_a=\bar s_a
  \qquad(a=1,\ldots,K).
\]
A feasible selector family exists by taking one common point of $\mathcal S$ in
every row. \qed

Lemma~\ref{lem:aipw-mean-coherence} permits a different
$\mathcal S$-valued selector at each outcome support point in each action row;
after optimizing the baseline, the rows are linked only through the displayed
own-coordinate mean constraints.

For the theorem, $\mathcal S=\mathcal C_b$. Since
$\mathbf1^\top\alpha_b(h)=\sum_j b_j$ for every history $h$,
$\mathcal S$ is constant-sum.

\paragraph{Sufficiency.}
Suppose first that the outcome marginals have finite support and let
$\{u_{ai}\}$ attain Lemma~\ref{lem:aipw-mean-coherence}. Express each $u_{ai}$
as a finite convex combination of extreme points and refine the row probability
space by this auxiliary randomization. This gives
$X^{(a)}\in\operatorname{ext}(\mathcal S)$ such that
\[
  \E\{X^{(a)}\mid Y(a)=y_{ai}\}=u_{ai}.
\]
The objective and row means are unchanged. Let $\bar X$ have the
behavior-weighted mixture law
\[
  \mathcal L(\bar X)=\sum_b\mu_b\mathcal L(X^{(b)}).
\]
The row constraints give
\[
  \E X^{(a)}_a
  =\sum_b\mu_b\E X^{(b)}_a
  =\E\bar X_a.
\]
If coordinate $a$ takes at most two values on
$\operatorname{ext}(\mathcal S)$, equality of these means implies
\[
  \mathcal L(X^{(a)}_a)=\mathcal L(\bar X_a).
\]
For each action, take from the row construction the conditional law of $Y(a)$
given $X^{(a)}_a$. Sample one common $\bar X$ from the displayed mixture law
and, conditional on $\bar X$, draw each potential outcome from the
corresponding kernel indexed by $\bar X_a$. Then the outcome marginals are the prescribed $F_a$ and
\[
  \E\{\bar X_aY(a)\}
  =\E\{X^{(a)}_aY(a)\}
  \qquad\text{for every }a.
\]
Since $\bar X\in\mathcal S$ a.s.,
\[
  h_{\mathcal S}(\mathbf Y)\ge\bar X^\top\mathbf Y.
\]
Thus
\[
  U_{\mathcal S}\ge A^+_{\mathcal S}.
\]
The outer bound proved above gives the reverse inequality, so equality holds
for finite-support marginals.

To pass from finite-support to general integrable marginals, note that
both upper functionals obey the same $W_1$ continuity bound:
\[
\begin{aligned}
  \left|
    U_{\mathcal S}(F_1,\ldots,F_K)
    - U_{\mathcal S}(\widetilde F_1,\ldots,\widetilde F_K)
  \right|
  &\le
  \sum_{a=1}^K
  \left(\sup_{u\in\mathcal S}|u_a|\right)
  W_1(F_a,\widetilde F_a),\\[4pt]
  \left|
    A^+_{\mathcal S}(F_1,\ldots,F_K)
    - A^+_{\mathcal S}(\widetilde F_1,\ldots,\widetilde F_K)
  \right|
  &\le
  \sum_{a=1}^K
  \left(\sup_{u\in\mathcal S}|u_a|\right)
  W_1(F_a,\widetilde F_a).
\end{aligned}
\]
The first bound follows by coupling the outcome coordinates in the sharp
problem, and the second holds uniformly over the AIPW baseline $\xi$.
Finite-support laws are dense in $W_1$ among integrable laws, so the
finite-support equality extends to all integrable marginals. Sign reversal
gives the lower endpoint and completes sufficiency.

\paragraph{Necessity.}
Suppose some coordinate takes at least three values on
$\operatorname{ext}(\mathcal S)$. The first lemma finds nearby points with
the same value of that coordinate but different attainable row payoff; the
second redistributes the resulting displacement across the remaining rows
without changing their own coordinates.

\begin{lemma}[Three-level spread separation]
\label{lem:local-spread-separation}
Suppose coordinate $a$ takes at least three distinct values across
$\operatorname{ext}(\mathcal S)$. Then there exist $p\in(0,1)$ and a compact
line segment $I\subset\operatorname{relint}(\mathcal S)$ on which $s_a$ is
constant. For $s\in\mathcal S$, define
\[
  \Phi_{a,p}(s)
  =
  \max\left\{
      p x^1_a:
      x^0,x^1\in\mathcal S,
      (1-p)x^0+px^1=s
    \right\},
\]
the largest weighted $a$-coordinate of the second point among two-point
decompositions of $s$ with weights $(1-p,p)$. Then $\Phi_{a,p}$ is finite,
concave, upper semicontinuous on $\mathcal S$, and nonconstant on $I$. Consequently, for every $\varepsilon>0$ there exist
$s_0,s_1\in I$ such that
\[
  \|s_1-s_0\|<\varepsilon,
  \qquad
  \Phi_{a,p}(s_1)>\Phi_{a,p}(s_0).
\]
\end{lemma}

\paragraph{Proof.}
Choose extreme points $v^-,v^0,v^+$ with
\[
  v^-_a=\ell<m=v^0_a<u=v^+_a,
  \qquad
  p=\frac{m-\ell}{u-\ell}.
\]
Set
\[
  s^\star=(1-p)v^-+pv^+.
\]
Then $s^\star_a=m$. Since $v^0$ is extreme, the only representation
\[
  (1-p)x^0+px^1=v^0
\]
with $x^0,x^1\in\mathcal S$ is $x^0=x^1=v^0$. Hence
\[
  \Phi_{a,p}(v^0)=pm,
  \qquad
  \Phi_{a,p}(s^\star)\ge pu>pm.
\]
Concavity follows by mixing feasible pairs. For upper semicontinuity, if
$s_n\to s$, a convergent subsequence of maximizing pairs has a feasible limit at
$s$, so
\[
  \limsup_n\Phi_{a,p}(s_n)\le\Phi_{a,p}(s).
\]

Choose $r\in\operatorname{relint}(\mathcal S)$ and mix both $v^0$ and
$s^\star$ with $r$ using the same weight $t<1$. For $t$ sufficiently close to
one, the two resulting relative-interior points have the same $a$ coordinate
and retain the strict ordering of their $\Phi_{a,p}$ values. Let $I$ be the
segment between them. Partitioning $I$ into sufficiently many equal pieces
produces adjacent $s_0,s_1$ with arbitrarily small distance and
$\Phi_{a,p}(s_1)>\Phi_{a,p}(s_0)$. \qed

\begin{lemma}[Constant-sum row balancing]
\label{lem:constant-sum-row-balancing}
Let $T$ be a linear subspace of
$\{d\in\mathbb R^K:\mathbf1^\top d=0\}$ with $\dim(T)\ge2$. Fix action $a$ and
define
\[
  K_b=\{d\in T:d_b=0\},
  \qquad b\ne a.
\]
Then
\[
  \sum_{b\ne a}K_b=T.
\]
Moreover, there is a finite constant $C_T$ such that every $d\in T$ has a
decomposition
\[
  d=\sum_{b\ne a}w_b,
  \qquad
  w_b\in K_b,
  \qquad
  \sum_{b\ne a}\|w_b\|\le C_T\|d\|.
\]
\end{lemma}

\paragraph{Proof.}
Let $f_b:T\to\mathbb R$ be the coordinate functional $f_b(d)=d_b$, so
$K_b=\ker(f_b)$. If $f_b$ vanishes identically for some $b\ne a$, then
$K_b=T$ and the claim is immediate. Otherwise each $K_b$ is a hyperplane in
$T$. If two such hyperplanes $K_b$ and $K_c$ are distinct, then
$K_b+K_c=T$, so again the claim follows.

It remains only to rule out the case in which all $K_b$, $b\ne a$, are the
same hyperplane. Nonzero linear functionals with the same kernel are
proportional, so all $f_b$, $b\ne a$, would then be proportional to one
nonzero functional on $T$. Because $T\subseteq\{d:\mathbf1^\top d=0\}$,
\[
  f_a=-\sum_{b\ne a}f_b
\]
on $T$, so $f_a$ is proportional to the same functional. Hence every
coordinate functional on $T$ is proportional to one scalar functional, which
would force the inclusion $T\hookrightarrow\mathbb R^K$ to have rank at
most one. This contradicts $\dim(T)\ge2$. Therefore
\[
  \sum_{b\ne a}K_b=T.
\]
The corresponding linear map from $\bigoplus_{b\ne a}K_b$ onto $T$ is
surjective. In finite dimension it has a bounded linear right inverse, giving
the stated constant. \qed

We now construct the finite-support witness. Assume $\mathcal S$ is not coordinatewise two-level. Choose action $a$ from
Lemma~\ref{lem:local-spread-separation} and let
\[
  T=\operatorname{lin}(\mathcal S-\mathcal S).
\]
Because $\mathcal S$ is constant-sum,
$T\subseteq\{d:\mathbf1^\top d=0\}$. A compact convex set of affine
dimension at most one is a point or line segment and is two-level, so
$\dim(T)\ge2$.

The segment $I$ lies in $\operatorname{relint}(\mathcal S)$, so there exists
$\eta>0$ such that
\[
  s+d\in\mathcal S
\]
whenever $s\in I$, $d\in T$, and $\|d\|\le\eta$. By
Lemma~\ref{lem:constant-sum-row-balancing}, choose $s_0,s_1\in I$ with
$d=s_1-s_0$ sufficiently small that
\[
  d=\sum_{b\ne a}w_b,
  \qquad
  w_b\in K_b,
  \qquad
  \left\|\frac{\mu_a}{\mu_b}w_b\right\|\le\eta
  \qquad(b\ne a),
\]
and
\[
  \Phi_{a,p}(s_1)>\Phi_{a,p}(s_0).
\]
Define
\[
  s^{(a)}=s_1,
  \qquad
  s^{(b)}=s_0-\frac{\mu_a}{\mu_b}w_b,
  \quad b\ne a.
\]
Then
\[
  s^{(b)}\in\mathcal S,
  \qquad
  \sum_b\mu_b s^{(b)}=s_0,
  \qquad
  s^{(b)}_b=(s_0)_b
  \quad\text{for every }b.
\]

Since $s_0\in\operatorname{relint}(\mathcal S)$ and $\Phi_{a,p}$ is finite and
concave, take a relative supergradient represented by $g\in\mathbb R^K$ and put
$c=-g$. Let $B_p\sim\operatorname{Bernoulli}(p)$ and define
\[
  Y(a)=c_a+B_p,
  \qquad
  Y(b)=c_b,
  \quad b\ne a.
\]
For the sharp common-selector problem,
\[
  U_{\mathcal S}
  =
  \max_{s\in\mathcal S}
  \{\Phi_{a,p}(s)+c^\top s\}
  =
  \Phi_{a,p}(s_0)+c^\top s_0.
\]
For the AIPW primal, choose $x^0,x^1$ attaining $\Phi_{a,p}(s_1)$ in row $a$
and use the single selector $s^{(b)}$ in every deterministic row $b\ne a$.
The row constraints hold, and the resulting value is
\[
  \Phi_{a,p}(s_1)+c^\top s_0
  >
  \Phi_{a,p}(s_0)+c^\top s_0
  =U_{\mathcal S}.
\]
Thus $A^+_{\mathcal S}>U_{\mathcal S}$ for finite-support outcome marginals.

Finally, write
$s_{\mathcal S}:=\mathbf1^\top u=\sum_j b_j$ for the common coordinate sum
of $u\in\mathcal S$. A common positive affine transformation of all outcomes
sends
\[
  U_{\mathcal S}
  \mapsto
  \alpha U_{\mathcal S}+\beta s_{\mathcal S},
  \qquad
  A^+_{\mathcal S}
  \mapsto
  \alpha A^+_{\mathcal S}+\beta s_{\mathcal S}.
\]
Choosing $\alpha>0$ and $\beta$ places the finite witness in any prescribed
bounded interval of positive length while preserving the strict inequality. Sign reversal gives a
lower-endpoint witness. This proves necessity and completes
Theorem~\ref{thm:aipw-universal-sharpness}. \qed

\endgroup

\section{Empirical design and supplementary diagnostics}

This appendix documents the empirical construction, provenance, and diagnostics
supporting Section~\ref{sec:empirical}.

\begingroup
\let\section\subsection
\let\subsection\subsubsection
\let\subsubsection\paragraph
\section{KuaiRec: construction and analysis}
\label{app:kuairec-empirical}

\subsection{Tasks, contract, and release refinements}

We use the public KuaiRec small matrix and user features
\citep{gao2022kuairec}. Users are split once by user-index parity into design
and evaluation groups. Video selection, policy construction, and refinement
fitting use the design group; evaluation-group outcomes are withheld from these
choices and from the matched-entropy worst-case ambiguity comparison. Candidate videos are
mutually disjoint across tasks, yielding 24 tasks and 144 distinct videos.

The task grid crosses
\[
\begin{aligned}
K&\in\{4,6,8\},\qquad
\text{rate regime}\in\{\text{low, central, high, mixed}\},\\
\text{composition}&\in\{\text{concentrated, diverse}\}.
\end{aligned}
\]
Low, central, and high select videos near prespecified lower, middle, and upper
design-user completion-rate regions; mixed combines these regions within one
action set. Concentrated tasks draw actions sharing the same first-listed item tag, whereas diverse tasks use distinct primary categories.

Behavior is uniform over the \(K\) actions. The base release \(G\) retains the
two coarse pre-action activity variables
\texttt{user\_active\_degree} and \texttt{is\_lowactive\_period}. A regularized
logistic model is fit for each candidate action on design users only. The
resulting scores define four target policies: global, which uses the action with
the highest design-group mean for every user; greedy, which uses the highest
predicted-score action for each user; top-two with masses \(0.75/0.25\); and
softmax with temperature \(0.25\). The declared evaluation contract contains
the four policy values and all six pairwise contrasts.

Each refinement partitions every fiber of the same base release into at most
\(m\) categories, with
\[
  m\in\{1,2,3,4,6,8,16,32,64\}.
\]
The five refinement rules are:
\begin{itemize}[leftmargin=*]
\item \textbf{Feature}: deterministic farthest-first refinement in standardized
pre-action feature space.
\item \textbf{Random}: random hashing with three prespecified seeds, summarized
by the taskwise median.
\item \textbf{Rank}: farthest-first refinement of the per-action score-rank
representation.
\item \textbf{Policy-vector}: farthest-first refinement of the four target-policy
vectors.
\item \textbf{Repair}: Algorithm~\ref{alg:finite-data-repair} with the
Section~\ref{sec:adequacy} action-coefficient distance. Representative histories
define the categories, and the resulting partition is evaluated under the
$d$-based objective without changing those assignments.
\end{itemize}

\subsection{Matched-entropy release-level analysis}

For each task and refinement, we measure worst-case ambiguity on evaluation
users without using their outcomes. For each evaluation direction, the
refined-cell \(\ell_1\) radii are averaged over evaluation users and divided by
the corresponding base-release average, with ratio zero when the base average
is zero. Figure~\ref{fig:kuairec-matched-information} reports $100$ times one minus
the largest of these ratios over the ten directions.

A common category limit \(m\) need not give refinements the same conditional
entropy. Each task--refinement path is therefore parameterized by
\(H(\Pi\mid G)\), the observed conditional entropy of refinement-category
membership given the base release. Duplicate entropy values are checked for consistent ambiguity ratios before
the trajectories are linearly interpolated on the common entropy support shared
by all 24 tasks and five refinements. Figure~\ref{fig:kuairec-matched-information} reports the
midpoint and endpoint of this common support.

\subsection{Realized sharp analysis and validity checks}

Evaluation-group outcomes are then used without changing the tasks, policies, or
fitted refinements. For each task, refinement, and \(m\), we compute the sharp
interval for each of the six pairwise policy contrasts. Each refined sharp width
is divided by its unrepaired-base width, and the six contrast-wise ratios are
averaged within task. Figure~\ref{fig:kuairec-consequence} reports the median and
interquartile range across tasks of \(100\) times one minus this mean ratio.
Sign certification is evaluated on the same \(24\times6=144\)
task--contrast pairs and requires the sharp interval to exclude zero.

As validity checks, complete-data values for all evaluated directions lie
inside their sharp intervals to numerical tolerance, and realized category
counts and conditional entropies are recomputed from the applied categories.

\endgroup

\begingroup
\let\section\subsection
\let\subsection\subsubsection
\let\subsubsection\paragraph
\section{HeartSteps: provenance and coarsening}

The public HeartSteps V1 interface contains both a prior-40-minute step
aggregate and a minute-level step stream \citep{klasnja2019heartsteps}.
Separately, a later HeartSteps design uses fewer than 150 steps in the preceding
40 minutes as a threshold \citep{liao2018timing}. We use these records to study
the preservation question in Section~\ref{sec:empirical-heartsteps}, without
claiming that the native V1 policy used the later threshold.

\subsection{Data construction and timing convention}

The analysis uses the public HeartSteps V1 \texttt{suggestions.csv} and
\texttt{jbsteps.csv} files. A time point is included when the corresponding
source decision is available, has an observed timestamp and native
prior-40-minute aggregate, and its user appears in the public minute-level step
stream. These criteria yield 6,590 time points. The timestamps serve only as
anchors for the preceding-40-minute records; actions and post-action outcomes
are not used.

The native prior-40-minute aggregate is
\texttt{jbsteps40pre.zero} when available and \texttt{jbsteps40pre} otherwise.
Flooring the recorded timestamp to the minute and reaggregating the public step
stream reproduces the recorded prior-40-minute count exactly on 99.71\% of time
points, with mean absolute discrepancy 0.19 steps. For the 150-step threshold,
the reconstruction and recorded aggregate agree on whether the count is below
150 at all 6,590 time points. Reaggregating at the exact recorded timestamp
instead changes whether the count is below 150 at 86 time points (1.31\%). Because the step stream is only
minute-level, neither reconstruction is treated as recovering sensor-level
within-minute ground truth.

\subsection{Coarsening and unresolved records}

After removing the recorded 40-minute aggregate, we deterministically group the
minute-level stream into \(w\in\{5,10,15,30\}\)-minute bins with one-minute
offset \(p\in\{0,\ldots,w-1\}\). A time point is unresolved when the same coarsened record can correspond to
minute-level histories whose 40-minute totals lie on opposite sides of the
150-step threshold. Table~\ref{tab:heartsteps-ablation} reports the ranges and
medians across offsets.

Adding back the same recorded 40-minute aggregate to each coarsened record
resolves the threshold distinction at every time point without restoring finer
temporal detail. The ablation does not evaluate native HeartSteps V1 policy
adequacy or claim that the later threshold motivated the original V1 recording
design.

\endgroup

\section{Positioning and scope}

This appendix locates the release problem relative to related OPE reduction,
partial-identification, and design formulations.

\subsection{Extended positioning and related formulations}
\label{app:extended-positioning}

\paragraph{OPE abstraction, quotienting, and target-oriented compression.}
The release problem fixes a deterministic pre-action map $R$ and an evaluation
contract $\mathcal E$, and asks whether the released law identifies every
value and comparison in the contract. Theorem~\ref{thm:estimand-quotient} gives the
exact criterion: the evaluation weights for every direction in $\mathcal E$
must be constant on each release fiber, and their joint level sets give the
coarsest exact release for the contract. State-abstraction and representation methods in OPE construct reduced
state or state--action descriptions tailored to evaluation
\citep{pavse2023rope,chaudhari2024arp,hao2024abstracted}. Conditional importance sampling uses conditional expectations of importance-sampling
ratios given selected information \citep{rowland2020conditional}, while
quotient constructions merge generation histories that are equivalent for
evaluation \citep{xie2026quotient}. Target-oriented compression
and decision-centric memory organize reduction around a downstream inferential
or decision target
\citep{chang2026target,zou2026decision}. These formulations share the principle
that retained information is judged relative to downstream use; for the release
problem, the criterion is identification of the evaluation contract from the
released law. This criterion can hold without reconstructing the fine-history
behavior policy. For example, if $\pi=\mu$, the evaluation weight is identically
one, so the corresponding value is identified by the release even when $\mu$
varies within a release fiber. Classical sufficiency theory relates information-preserving reductions in
dominated experiments to likelihood-ratio structure \citep{halmossavage1949},
while partial-sufficiency work connects target-specific reduction to
identification under partial observability \citep{oulhajmouchart2003}.
For a single target policy, the evaluation weight is the target-to-behavior
likelihood ratio on $(H,A)$; here the criterion is identification from the
released action--outcome law for a declared linear contract, including signed
comparisons. Classical comparison of experiments provides a framework for
comparing the informativeness of statistical experiments \citep{blackwell1953}.

\paragraph{Partial identification and optimal recovery under release loss.}
When a release is not exact for a direction $b$, the corresponding evaluation
is partially identified from the released law. Related work studies partial
identification under limited overlap or noisy covariates
\citep{khan2024overlap,guo2022noisy}, and decision ambiguity under degraded
observed signals \citep{shekhar2026signal}. In the release setting,
Theorem~\ref{thm:general-evaluation-radius} identifies the largest sharp width
over attainable $Q_g$ with $\Omega_g(b)$, a behavior-weighted within-fiber
evaluation-weight radius that allows behavior to vary within a release fiber.
Classical optimal recovery connects statistical estimation with minimax
approximation from indirect information \citep{donoho1994optimal}. The zero set of $\Omega_g(b)$ recovers exactness,
and Section~\ref{sec:adequacy} uses the same distortion as the
finite-category refinement objective.

\paragraph{Temporal observation and latent-process formulations.}
Work on temporal resampling and misalignment studies changes to temporal
observation structure \citep{tang2026offbeat,frost2026hidden}, while
confounded-POMDP formulations study latent decision-process structure
\citep{shi2022pomdp}. The release formulation
holds the fine-history space and the stated OPE model fixed and studies the
information retained by a deterministic pre-action interface.

\paragraph{Sharp identification and AIPW sharpness.}
For a fixed $Q_g$, Section~\ref{sec:transport} first characterizes when
$Q_g$ identifies each action-specific outcome distribution: exactly under the
within-fiber constancy condition that defines release-measurable behavior.
This condition is the finite-action analogue of a balancing-score reduction:
it makes $A \perp\!\!\!\perp H\mid G$, paralleling propensity-score and
general-treatment constructions \citep{rosenbaumrubin1983,imaivandyk2004}.
Proposition~\ref{prop:action-marginal-from-release} uses it as the exact
gate for identification of the action-specific outcome distributions from
$Q_g$. Under this condition, history variation is summarized by the
action-coefficient set, while the remaining uncertainty is how the identified
action-specific outcome distributions are coupled. Multi-marginal transport provides machinery for partial
identification from marginal information \citep{gao2025multimarginal}.
Computing the sharp interval for $V_b$ directly retains the joint restriction
that its component policy values come from the same underlying law; a policy
comparison can therefore have a narrower interval than one obtained from
separate value intervals. When behavior varies within a release fiber,
Appendix~\ref{app:general-behavior-sharp} gives the corresponding
behavior-weighted row-constraint formulation.

Classical doubly robust OPE provides the score underlying the fixed-coefficient
AIPW construction \citep{dudik2011doubly}.
Section~\ref{sec:contract-geometry-bridge} uses that
score to form an envelope over the action-coefficient set and compares the
envelope with the sharp interval. Theorem~\ref{thm:aipw-universal-sharpness}
characterizes universal equality: each action coordinate takes at most two
distinct values across the extreme points of the action-coefficient set.

\paragraph{Dataset support and logging-policy design.}
DataCOPE asks whether and to what extent a target policy can be evaluated from
a given logged dataset \citep{sun2024datacope}. Logging-policy design chooses
the behavior action probabilities used to collect OPE data
\citep{douglas2026logging}. In the finite-category refinement
problem, those probabilities remain fixed inputs and the design variable is a
finite-category refinement of the base release. These formulations concern
different parts of the OPE interface: dataset support, action allocation, and
retained pre-action information.

\subsection{Scope boundaries}
\label{app:scope-limitations}

The release mechanism studied here is a deterministic pre-action map.
Replacing it by a known stochastic channel changes how histories map to
released records and therefore changes both the identification and
finite-category refinement problems.

The policy-value problem considered has a single decision point, although $H$ may encode arbitrarily rich prior history. Multi-step sequential target-policy evaluation adds trajectory-level structure, such as state transitions, and can also involve hidden confounding or unsupported actions, which change the identification problem.

\end{document}